%% file: main.tex
\pdfoutput=1

\documentclass[11pt]{article}

\usepackage[final]{coling}

\usepackage{times}
\usepackage{latexsym}
\usepackage{url}
\usepackage{multirow}
\usepackage{enumitem}
\usepackage{adjustbox}
\usepackage{tcolorbox}
\usepackage{booktabs} 
\usepackage{amsfonts}
\usepackage[T1]{fontenc}
\usepackage[utf8]{inputenc}
\usepackage{microtype}
\usepackage{inconsolata}
\usepackage{graphicx}
\usepackage{xspace}

\setlist{leftmargin=*,nosep}

\newcommand{\named}{NyayaAnumana\xspace}
\newcommand{\namel}{INLegalLlama\xspace}

\title{\textsc{\named} and \textsc{\namel}: \\The Largest Indian Legal Judgment Prediction Dataset and Specialized Language Model for Enhanced Decision Analysis}

\author{Shubham Kumar Nigam$^{1}$ \qquad Balaramamahanthi Deepak Patnaik$^{1}$ \qquad  Shivam Mishra$^{1}$ \\ 
\textbf{Noel Shallum}$^{3}$ \qquad \textbf{Kripabandhu Ghosh}$^{2}$ \qquad \textbf{Arnab Bhattacharya}$^{1}$\\
$^{1}$ IIT Kanpur, India \quad
$^{2}$  IISER Kolkata, India \quad
$^{3}$ Symbiosis Law School Pune, India \\
\texttt{\{sknigam, deepak, shivammishra, arnabb\}@cse.iitk.ac.in} \\
\texttt{kripaghosh@iiserkol.ac.in,} \quad \texttt{noelshallum@gmail.com}
}

\date{}

\begin{document}
\maketitle


\input{abstract}

\input{intro}

\input{related_work}

\input{task_description}

\input{dataset}

\input{InLegalLlama_training}

\input{methodology}

\input{evaluation_metrics}

\input{results_and_analysis}

\input{conculsion}

\input{acknowledgement}

\input{limitation}

\input{ethics}
\newpage
\bibliography{anthology, custom}

\newpage
\appendix
\input{appendix}

\end{document}

%% file: abstract.tex
\begin{abstract}
The integration of artificial intelligence (AI) in legal judgment prediction (LJP) has the potential to transform the legal landscape, particularly in jurisdictions like India, where a significant backlog of cases burdens the legal system. This paper introduces \texttt{\named}, the largest and most diverse corpus of Indian legal cases compiled for LJP, encompassing a total of 7,02,945 preprocessed cases. \texttt{\named}, which combines the words ``Nyaya'' and ``Anumana'' that means ``judgment'' and ``inference'' respectively for most major Indian languages, includes a wide range of cases from the Supreme Court, High Courts, Tribunal Courts, District Courts, and Daily Orders and, thus, provides unparalleled diversity and coverage. Our dataset surpasses existing datasets like PredEx and ILDC, offering a comprehensive foundation for advanced AI research in the legal domain.
In addition to the dataset, we present \texttt{\namel}, a domain-specific generative large language model (LLM) tailored to the intricacies of the Indian legal system. It is developed through a two-phase training approach over a base LLaMa model. First, Indian legal documents are injected using continual pretraining. Second, task-specific supervised finetuning is done. This method allows the model to achieve a deeper understanding of legal contexts.
Our experiments demonstrate that incorporating diverse court data significantly boosts model accuracy, achieving approximately 90\% F1-score in prediction tasks. \texttt{\namel} not only improves prediction accuracy but also offers comprehensible explanations, addressing the need for explainability in AI-assisted legal decisions. 


\end{abstract}

%% file: intro.tex
\section{Introduction}
\begin{table*}[t]
\centering
\resizebox{\linewidth}{!}{%
\begin{tabular}{|l|c|c|r|r|l|}
\hline
\textbf{Corpus} &
  \textbf{Language} &
  \textbf{Jurisdiction} &
  \textbf{\# of Cases} &
  \begin{tabular}[c]{@{}c@{}}\textbf{Avg \# of} \\ \textbf{Tokens}\end{tabular} &
  \multicolumn{1}{c|}{\textbf{\begin{tabular}[c]{@{}c@{}}\# of labels w.r.t Subtask\end{tabular}}} \\ \hline
\begin{tabular}[c]{@{}c@{}}FCCR \cite{csulea2017exploring}\end{tabular} &
  French &
  France &
  1,26,865 &
  - &
  \begin{tabular}[c]{@{}l@{}}Court Decision(6 \& 8)\end{tabular} \\ \hline
\begin{tabular}[c]{@{}c@{}}CAIL \cite{xiao2018cail2018}\end{tabular} &
  Chinese &
  China &
  26,76,075 &
  - &
  \begin{tabular}[c]{@{}l@{}}Law Article (183)\\ Charge (202)\end{tabular} \\ \hline
\begin{tabular}[c]{@{}c@{}}ECHR \cite{chalkidis2019neural}\end{tabular} &
  English &
  Europe &
  11,478 &
  2,421 &
  \begin{tabular}[c]{@{}l@{}}Violation (2) \\ Law Article (66)\end{tabular} \\ \hline
\begin{tabular}[c]{@{}c@{}}ECHR \cite{chalkidis-etal-2021-paragraph}\end{tabular} &
  English &
  Europe &
  11,000 &
  - &
  \begin{tabular}[c]{@{}l@{}}Alleged Law Article (40)\\ Violation (2)\\ Law Article (40)\end{tabular} \\ \hline
\begin{tabular}[c]{@{}c@{}}SJP \cite{niklaus2021swiss}\end{tabular} &
  \begin{tabular}[c]{@{}c@{}}German\\ French\\ Italian\end{tabular} &
  Switzerland &
  \begin{tabular}[c]{@{}c@{}}49,883 \\ 31,094 \\ 4,292 \end{tabular} &
  850 &
  Court Decision (2) \\ \hline
\begin{tabular}[c]{@{}c@{}}ILDC \cite{malik-etal-2021-ildc}\end{tabular} &
  English &
  India &
  34,816 &
  3,231 &
  Court Decision (2) \\ \hline
\begin{tabular}[c]{@{}c@{}}HLDC \cite{kapoor-etal-2022-hldc}\end{tabular} &
  Hindi &
  India &
  3,40,280 &
  764 &
  Bail Prediction (2) \\ \hline
\begin{tabular}[c]{@{}c@{}}BCD \cite{lage2022predicting}\end{tabular} &
  Portuguese &
  Brazil &
  4,043 &
  119 &
  \begin{tabular}[c]{@{}l@{}}Court Decision (3)\end{tabular} \\ \hline
\begin{tabular}[c]{@{}c@{}}PredEx \cite{nigam2024legaljudgmentreimaginedpredex} \end{tabular} &
  English &
  India &
  15,222 &
  4,504 &
  \begin{tabular}[c]{@{}l@{}}Court Decision (2)\end{tabular} \\ \hline
\begin{tabular}[c]{@{}c@{}}\textbf{(Our dataset)} \textbf{\texttt{\named}}\end{tabular} &
  English &
  India &
  7,02,945 &
  2,061 &
  \begin{tabular}[c]{@{}l@{}}Court Decision (2 \& 3)\end{tabular} \\ \hline
\end{tabular}%
}
\caption{Comparative overview of widely used legal judgment prediction datasets. Entries marked with `-' denote unknown or unavailable information.}
\label{tab:old-data-stat}
\end{table*}

The integration of artificial intelligence (AI) in legal judgment prediction (LJP) has the potential to revolutionize the legal landscape, offering both challenges and opportunities. In India, where the legal system faces a significant backlog of lakhs of pending cases, the application of AI in LJP can be crucial for enhancing efficiency and accessibility. However, the complexity and diversity of legal cases present significant challenges in developing effective AI models. To address these challenges, we introduce \textbf{\texttt{\named}}, the largest and most diverse corpus of Indian legal cases compiled for LJP, covering various levels of the judiciary. The name ``NyayaAnumana'' is formed by a combination of the words
``Nyaya'' and ``Anumana'' that mean ``judgment''
and ``inference'' respectively for most major Indian languages. This name reflects the core focus of the dataset on legal judgments and their corresponding predictions, emphasizing its role in facilitating AI-driven insights within the legal domain. Our corpus stands out when compared to other popular corpora used in legal judgment prediction, surpassing them in terms of the number of cases, diversity of court levels, and comprehensive coverage of Indian legal proceedings, as shown in Table~\ref{tab:old-data-stat}. This richness and variety offer a unique opportunity to explore and predict legal judgments more accurately and nuanced than ever before. 

We develop \textbf{\texttt{\namel}}, a domain-specific generative large language model, tailored to the intricacies of the Indian legal domain. It is trained to enhance the accuracy of predictions of legal judgments and provide understandable explanations for these decisions. This dual approach caters to the needs of legal experts who seek not just accuracy but also rationale in AI-assisted decisions.

Our work is distinguished by several key contributions that mark significant advancements in the field of legal AI:
\begin{enumerate}
\item \emph{Largest Indian Legal Corpus for Judgment Prediction:} We introduce \texttt{\named}, the most extensive legal corpus in India for LJP, encompassing a wide range of courts and orders, ensuring diversity and comprehensive coverage in the dataset.
\item \emph{Generative Model for Prediction and Explanation:} Addressing the crucial need for explainability, we develop \texttt{\namel}, a generative LLM that not only predicts outcomes but also provides comprehensible explanations, enhancing the trustworthiness and utility of AI in legal decision-making.
\item \emph{Domain-Specific Transformer-Based Classifiers:} We fine-tuned several transformer-based models on \texttt{\named} to enhance prediction accuracy for the Indian legal domain. This includes testing model performance across different courts and hierarchical data to evaluate the impact of data inclusion.
\item \emph{Evaluation on Temporal Data:} We tested the model performance on temporal data, assessing its effectiveness on future or unseen data to ensure robustness and generalization capabilities over time.
\item \emph{Expert Evaluation and Validation:} We conducted a thorough evaluation using a Likert score scale to assess the effectiveness of our system. This evaluation utilized the PredEx test dataset \cite{nigam2024legaljudgmentreimaginedpredex} and ILDC\_expert \cite{malik-etal-2021-ildc}, offering crucial insights into how our AI models perform compared to human expert benchmarks.
\end{enumerate}

Our research aspires to deliver a sophisticated AI-driven system for legal judgment prediction and explanation, specifically designed for the Indian judicial system. This initiative not only represents a technological leap but also aims to tackle the critical issue of case backlog in India. We anticipate that our work will enhance the efficiency and transparency of the legal process and stimulate further research and development in the realm of AI-assisted legal technologies. 
For the sake of reproducibility, we have made the \texttt{\named} dataset and the code for our prediction and explanation models accessible via a GitHub link\footnote{\href{https://github.com/ShubhamKumarNigam/NyayaAnumana-and-INLegalLlama}{GitHub NyayaAnumana-and-INLegalLlama}}. 

%% file: related_work.tex
\section{Related Work}

The field of Legal Judgment Prediction (LJP) through AI has progressed significantly over the last few years. Traditionally the domain of legal experts, LJP systems offer potential benefits for practitioners and the public, particularly in managing overwhelming caseloads in various jurisdictions.
Foundational studies by \cite{aletras2016predicting}, \cite{chalkidis2019neural}, and \cite{feng2021recommending} established the methodologies for LJP and highlighted the critical need for explainability in AI-generated predictions. Benchmark datasets such as CAIL2018 \cite{xiao2018cail2018, zhong-etal-2020-nlp}, ECHR-CASES \cite{chalkidis2019neural, aletras2016predicting, medvedeva2020using}, and others have propelled research by providing a foundation for model evaluation. Despite these advances, challenges persist in achieving machine performance comparable to human expertise.


  

In the Indian context, significant contributions include PredEx \cite{nigam2024legaljudgmentreimaginedpredex} and ILDC \cite{malik-etal-2021-ildc}, which emphasize the importance of explainability in AI systems for legal reasoning. Fact-based judgment prediction has gained prominence as a realistic approach to LJP, focusing on predictions derived from case facts rather than full case judgments. \citet{nigam-etal-2024-rethinking} and \citet{10.1145/3632754.3632765} explore LJP based on facts, arguing that this approach better simulates real-world scenarios. \citet{nigam2022nigam, nigam-etal-2023-nonet, nigam2023legal, malik2021semantic, ganguly2023legal, ghosh2023report} underscore the growing role of AI in Indian legal applications, demonstrating the utility of advanced models like LLMs in addressing the unique challenges of this domain. Furthermore, \citet{tiwari2024aalap} and \citet{huang2023lawyer} illustrate the potential of AI assistants in legal and paralegal functions, showcasing the applicability of AI in streamlining legal processes and improving access to justice.
Cross-jurisdictional research, such as \cite{zhao-etal-2018-learning}, demonstrates the adaptability of LJP models across different legal systems and languages. The multilingual aspects of LJP are addressed by studies such as \cite{kapoor-etal-2022-hldc}, which introduces a corpus for Hindi legal documents, and \cite{niklaus2021swiss}, which examines the Swiss legal system.

%% file: task_description.tex
\section{Task Description}
Our research focuses on advancing the Legal Judgment Prediction (LJP) task, which encompasses two primary components: Prediction and Explanation. These components are executed sequentially to address the crucial needs of predicting legal judgments and providing justifications for these predictions. Figure \ref{fig:task-framework} in the Appendix provides a visual overview of the LJP framework utilized in our study.

\noindent
\textbf{Prediction Task:} The LJP task's core objective is to predict a legal case's outcome based on the case proceedings. Unlike previous studies primarily focusing on binary classification (acceptance or rejection), our study also classifies cases with partially accepted outcomes. Given a document \(D\), the task is to predict the decision \(y \in \{0, 1, 2\}\), where `0' denotes the rejection of all appeals by the appellant, `1' represents the acceptance of all appeals, and `2' indicates partial acceptance of the appeals.

\noindent
\textbf{Explanation Task:} The second component of the LJP task involves explaining the model's predicted decision. To address the need for explainability, we developed \texttt{\namel}, a generative LLM. Initially, the model was trained on case text to incorporate legal knowledge. Following this, we fine-tuned the model in a supervised manner to enhance its capabilities in both prediction and explanation, thereby increasing the reliability and usefulness of AI in legal decision-making.

%% file: dataset.tex
\section{Dataset}

\begin{table}[t]
\centering
\resizebox{\columnwidth}{!}{%
\begin{tabular}{lrrrr}
\hline
\multirow{2}{*}{\textbf{Metric}} & \multicolumn{2}{c}{\textbf{Train}} & \multirow{2}{*}{\textbf{Validation}} & \multirow{2}{*}{\textbf{Test}} \\
\cmidrule{2-3}
& \textbf{multi} & \textbf{single} &         &         \\ \hline
\multicolumn{5}{c}{\textbf{SCI}}                                             \\ \hline
\#Documents      & 35,942          & 20,712           & 2,960    & 5,919    \\
Avg \#words  & 2,738        & 2,734         & 2,638 & 2,731 \\
Acceptance(\%) & 59.37          & 50.79           & 50.74   & 50.87   \\ \hline
\multicolumn{5}{c}{\textbf{SCI + HCs}}                      \\
\hline
\#Documents       & 4,84,725         & 2,83,457          & 40,495   & 80,988   \\
Avg \#words  & 2,168        & 2,115         & 2,096 & 2,107 \\
Acceptance(\%) & 55.75          & 51.94           & 48.22   & 48.25   \\  \hline
\multicolumn{5}{c}{\textbf{SCI + HCs + Tribunals}} \\
\hline
\#Documents       & 6,48,356         & 3,68,639        & 52,663   & 1,05,326  \\
Avg \#words  & 2,081        & 1,999         & 2,003 & 1,989 \\
Acceptance(\%) & 58.27          & 49.48           & 49.49   & 49.36   \\ \hline
\multicolumn{5}{c}{\textbf{SCI + HCs + Tribunals + Daily Orders and District Courts}}    \\
\hline
\#Documents       & 7,02,945         & 4,01,412         & 57,345   & 1,14,690  \\
Avg \#words  & 2,061        & 1,985         & 1,987 & 1,976 \\
Acceptance(\%) & 57.98          & 49.30           & 49.34   & 49.04  
\\ \hline
\end{tabular}%
}
\caption{Data statistics across different courts for evaluating model performance on binary classification.}
\label{tab:binary-stats}
\end{table}

\begin{table}[t]
\centering
\resizebox{\columnwidth}{!}{%
\begin{tabular}{lrrr}
\hline
\textbf{Metric}  & \textbf{Train}   & \textbf{Validation}   & \textbf{Test}  
\\ \hline
\multicolumn{4}{c}{\textbf{SCI}}     
\\ \hline
\#Documents               & 26,823            & 3,833                 & 7,665           \\
Avg \#words               & 2,777          & 2,779               & 2,783        \\
Clear acceptance(\%)      & 41.95            & 41.22                 & 41.44          \\
Partial acceptance (\%)   & 17.19            & 18.03                 & 17.59          \\ \hline
\multicolumn{4}{c}{\textbf{SCI + HCs}}                      \\
\hline
\#Documents               & 3,40,972           & 48,711                 & 97,421          \\
Avg \#words               & 2,173          & 2,166               & 2,180        \\ 
Clear acceptance(\%)      & 54.31            & 54.46                 & 54.56          \\
Partial acceptance (\%)   & 1.38             & 1.29                  & 1.34           \\ \hline
\multicolumn{4}{c}{\textbf{SCI + HCs + Tribunals}}                                                            \\
\hline
\#Documents               & 4,55,514           & 65,074                 & 1,30,147         \\
Avg \#words               & 2,085          & 2,082               & 2,089        \\
Clear acceptance(\%)      & 57.23            & 57.55                 & 57.10          \\
Partial acceptance (\%)   & 1.03             & 0.96                  & 1.03           \\ \hline
\multicolumn{4}{c}{\textbf{SCI + HCs + Tribunals + Daily Orders and District Courts}}    \\
\hline
\#Documents               & 4,93,726           & 70,533                 & 1,41,065         \\
Avg \#words               & 2,061          & 2,068               & 2,081        \\
Clear acceptance(\%)      & 57.02            & 56.83                 & 57.14          \\
Partial acceptance (\%)   & 0.94             & 0.97                  & 0.95   \\ \hline       
\end{tabular}%
}
\caption{Data statistics across different courts for evaluating model performance on ternary classification.}
\label{tab:ternary-stats}
\end{table}

We introduce the dataset \textbf{\texttt{\named}}, which is the largest and most diverse corpus of Indian legal cases ever compiled, covering judgments from the Supreme Court, High Courts, Tribunal courts, District courts, and Daily orders. This comprehensive dataset offers unparalleled diversity and coverage, addressing existing gaps and providing a rich foundation for advanced legal AI research.

\subsection{Dataset Compilation}

The dataset compilation process involved gathering a corpus of 22,82,137 Indian court case proceedings up to April 2024. We utilized the IndianKanoon website\footnote{\url{https://indiankanoon.org/}}, a well-known legal search engine, to collect these documents. This source is widely recognized for its comprehensive database of Indian legal documents, making it an invaluable resource for our dataset.

\subsection{Data Statistics} 

The \texttt{\named} dataset exhibits extensive data statistics, vital for understanding the dataset's scope and characteristics. The overall dataset is divided into `multi' and `single' categories based on the nature of the decisions. Table~\ref{tab:binary-stats} and Table~\ref{tab:ternary-stats} provide detailed statistics, including the number of documents, the average number of tokens, and the distribution of clear and partial acceptance decisions across the dataset. This dataset is also analyzed on a court-wise basis, providing insights into the unique characteristics of cases from different courts.
Figure~\ref{fig:court-distribution} in the Appendix shows the distribution of cases in different courts in percentage. This breakdown helps in understanding the diversity within the dataset and the varying complexities associated with different court levels.


\subsubsection{Injecting Legal Knowledge}
To address the deficiency of legal knowledge in the base LLaMa model, we employed a \emph{continued pretraining (CPT)} approach using a comprehensive Indian legal corpus. Due to resource constraints, we used a subset of the full \texttt{\named} dataset for this pretraining phase. This subset includes preprocessed data comprising 38,321 cases from the Supreme Court of India (SCI) and a randomly selected 1,00,000 cases from various High Courts. These choices were made to balance computational feasibility and the inclusion of diverse and representative legal cases. This extensive yet manageable training corpus was essential for embedding domain-specific legal knowledge into the model. Additionally, the validation dataset consisted of 12,239 documents sourced from both SCI and High Courts, ensuring that the model was rigorously tested and fine-tuned for the nuances of the Indian legal system. By focusing on a strategically chosen subset, we aimed to maximize the impact of the training within the available computational resources. This approach enhances the model's understanding and applicability within the Indian legal framework, thereby improving its predictive capabilities and relevance in legal tasks.

\subsubsection{Learning Reasoning Skills}
To equip the model with the necessary reasoning capabilities for solving prediction and explanation problems, we conducted \emph{supervised fine-tuning (SFT)} using selected data from downstream tasks on the PredEx training dataset \cite{nigam2024legaljudgmentreimaginedpredex}, which consists of 12,178 cases accompanied by corresponding case decisions and explanations annotated by legal experts. By performing SFT on this dataset, we aimed to enhance its reasoning skills and ability to comprehend and apply legal principles effectively. This targeted fine-tuning approach helps bridge the gap between the model's general knowledge and the specific requirements of legal reasoning tasks, ultimately improving its performance in real-world legal scenarios.

\subsubsection{Prediction Task}

For prediction, we split the \texttt{\named} single dataset into training, validation, and test sets in the ratio 70:10:20. A key component of our research involved comparing the performance of models trained on \texttt{\named} with those trained on the ILDC 2021 \cite{malik-etal-2021-ildc} test dataset. This comparison is crucial for benchmarking our models and understanding their efficacy compared to established datasets in the field. By testing against ILDC 2021, we aim to evaluate the improvements in prediction accuracy and model robustness that \texttt{\named} offers, showcasing its contribution to the evolving landscape of legal AI in India.
We also tested the model performance on temporal data, assessing its effectiveness on future or unseen data from January 2020 to April 2024 to ensure robustness and generalization capabilities over time, as detailed in Table \ref{tab:temporal-test-cases}.


\begin{table}[t]
\centering
\resizebox{\columnwidth}{!}{%
\begin{tabular}{lrrrr}
\hline
\textbf{Metric}  &
  \textbf{SCI} &
  \textbf{HCs} &
  \textbf{Tribunals} &
  \textbf{\begin{tabular}[c]{@{}c@{}}Daily Orders and\\ District Courts\end{tabular}} \\ \hline
\textbf{\#Documents}      & 1,812    & 29,216   & 19,034   & 17,002   \\ 
\textbf{Avg \#words}  & 4,470 & 2,671 & 2,927 & 1,306 \\ 
\textbf{Acceptance (\%)} & 67.77   & 53.18   & 47.42   & 62.40   \\ \hline
\end{tabular}%
}
\caption{Statistics of temporal test data (January 2020 -- April 2024) across different courts for evaluation.}
\label{tab:temporal-test-cases}
\end{table}

\subsubsection{Prediction with Explanation Task}
For this task, we used the PredEx 2024 test dataset \cite{nigam2024legaljudgmentreimaginedpredex}, which includes 3,044 balanced cases. The balanced nature of the test set is particularly important for maintaining the validity of our experiments and ensuring the reliability and generalizability of our model's performance.


%% file: InLegalLlama_training.tex
\section{Model Training: \namel}

\subsection{Injecting Legal Knowledge}

To address the limitations of legal knowledge inherent in the base LLaMa model, we adopted a continued pretraining (CPT) strategy utilizing a comprehensive Indian legal corpus. For this purpose, we selected the LLaMa2-7B architecture \cite{touvron2023llama}, which features a substantial context length of 2K, allowing for effective handling of legal texts. This choice facilitates a direct comparison with previous state-of-the-art results on the PredEx dataset \cite{nigam2024legaljudgmentreimaginedpredex}. This approach significantly enhances the model's understanding and relevance within the Indian legal framework, thereby improving its predictive capabilities and application in legal tasks.

\subsection{Learning Reasoning Skills}

To further develop the model's reasoning skills, particularly for legal prediction and explanation tasks, we conducted supervised finetuning (SFT) using data from specific downstream tasks. This dataset includes case decisions along with their corresponding explanations, all annotated by legal experts. The fine-tuning process was crucial for enhancing the model's ability to understand and apply legal principles effectively, bridging the gap between general knowledge and the specialized requirements of legal reasoning tasks. This focused fine-tuning was pivotal in improving the model's performance in real-world legal scenarios.

However, the fine-tuning of such models typically demands substantial computational resources and extensive training data. Given the constraints of limited computational power and the specific nature of our legal task-related dataset, we prioritized efficient training methods to optimize both computational costs and data usage. We employed parameter-efficient tuning techniques, such as the Low-Rank Adaptation (LoRA) method \cite{hu2021lora}, to fine-tune the LLaMa-2 7B model. This approach enabled us to maximize the utility of available data and minimize the need for extensive computational resources, ensuring a cost-effective yet powerful fine-tuning process for developing \texttt{\namel}.

\begin{figure*}[t] 
    \centering 
    \includegraphics[width= \linewidth]{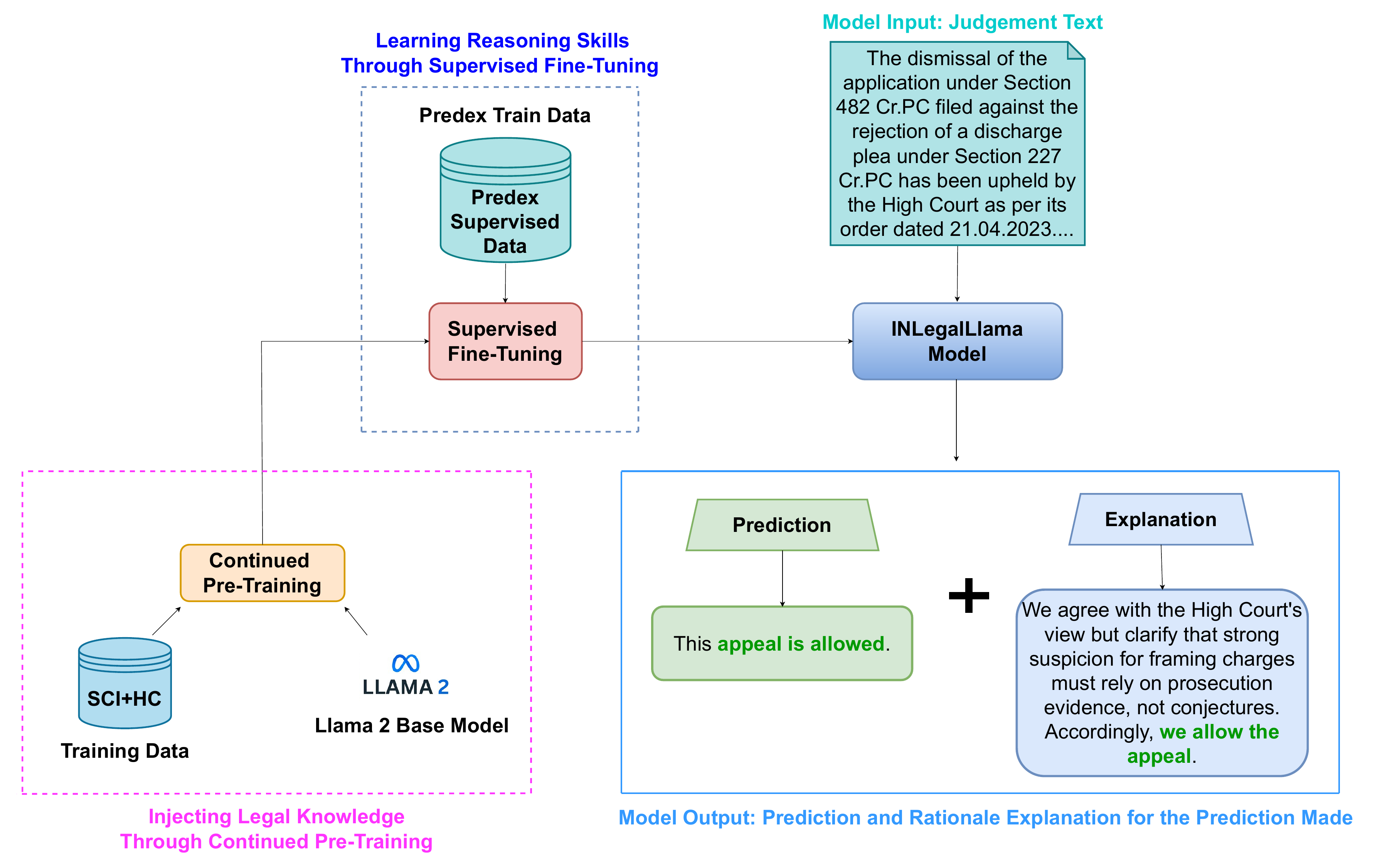} 
    \caption{\namel flow diagram}
    \label{fig:InLegalLlama-Flow-Diagram} 
\end{figure*}

%% file: methodology.tex
\section{Methodology}
\subsection{Judgment Prediction}
The judgment prediction task involves both binary classification (e.g., favoring or opposing a party) and ternary classification (e.g., fully accepted, partially accepted, or rejected). We employ two distinct approaches for this purpose: Language Model-based and LLM-based strategies. These approaches are carefully designed to handle the complexity and diversity of Indian legal documents, ensuring robust performance across various scenarios.

\subsubsection{Language Model based}
In our approach, we utilized several language models, including InLegalBERT, InCaseLaw \cite{paul-2022-pretraining}, and XLNet (large) \cite{yang2019xlnet}, as baselines for binary and ternary classification. Due to the length constraints of complete judgments, which exceed the token capacity of these models, we adopted a chunking strategy. Each document was divided into 512-token chunks using a moving window approach with a 100-token overlap to preserve textual context.

\subsubsection{Large Language Model based}
To utilize LLMs in prediction, we employed two strategies: one involving only prediction instructions and the other prediction with explanation instructions. We used two methods to get predictions from \texttt{\namel} after CPT and CPT followed by SFT. We followed the prompts and instruction-tuning approaches published by \cite{vats-etal-2023-llms} and \cite{nigam2024legaljudgmentreimaginedpredex} in a few-shot setup, and used the PredEx training data for instruction-tuning.

\subsection{Judgment Prediction with Explanation}
For this task, we employed the same LLMs with settings similar to the Judgment Prediction task, but with modified instructions focusing on both prediction and explanation.

\subsection{Prompts Used}
For both task inferences we utilized prompts from \cite{vats-etal-2023-llms}. These prompts, which include a case description and a gold standard prediction label, guide the LLM to generate judicial decisions. The details of these prompts can be found in Table \ref{tab:judgment_prediction_prompts_few} in the Appendix. In addition, for instruction tuning, we adopted prompts from \cite{nigam2024legaljudgmentreimaginedpredex} for prediction tasks, as listed in Table \ref{tab:judgment_prediction_prompts_zero} in the Appendix. The relevant details and examples of the generated predictions and explanations are also available in the referenced tables in the Appendix.

\subsection{Instruction-Set}
For both judgment prediction and explanation, we used 16 instruction sets correspondingly published by \cite{nigam2024legaljudgmentreimaginedpredex}. For a comprehensive view of all instruction sets which was randomly given to the model for tuning, we have included the full list in Table~\ref{Instruction-sets} in the Appendix of this paper.

%% file: evaluation_metrics.tex
\section{Evaluation Metrics}
\label{sec:performance_metrics}

In this study, we employed a comprehensive set of evaluation metrics to assess the performance of our models on the \named judgment prediction and PredEx explanation test datasets. We report Macro Precision, Macro Recall, Macro F1, and Accuracy for judgment prediction, and we use both quantitative and qualitative methods to evaluate the quality of explanations generated by the model.

\begin{enumerate}
    \item \textbf{Lexical-based Evaluation:} We utilized standard lexical similarity metrics, including Rouge scores (Rouge-1, Rouge-2, and Rouge-L) \cite{lin-2004-rouge}, BLEU \cite{papineni-etal-2002-bleu}, and METEOR \cite{banerjee-lavie-2005-meteor}. These metrics measure the overlap and order of words between the generated explanations and the reference texts, providing a quantitative assessment of the lexical accuracy of the model outputs.

    \item \textbf{Semantic Similarity-based Evaluation:} To capture the semantic quality of the generated explanations, we employed BERTScore \cite{BERTScore}, which measures the semantic similarity between the generated text and the reference explanations. Additionally, we used BLANC \cite{blanc}, a metric that estimates the quality of generated text without a gold standard, to evaluate the model's ability to produce semantically meaningful and contextually relevant explanations.
    
    \item \textbf{Expert Evaluation:} Human evaluation was a critical component of our assessment framework. Legal experts reviewed the explanations generated by the models, rating them on a 1–5 Likert scale based on criteria such as accuracy, relevance, and completeness. A rating of 1 indicates that the information is irrelevant, while a rating of 5 signifies that the explanation is superior to the expert's own explanation. The full description of the rating scores can be found in Appendix \ref{section:rating-score-description} - Rating Score Description, which is adapted from \cite{nigam2024legaljudgmentreimaginedpredex}.

\end{enumerate}

%% file: results_and_analysis.tex
\section{Results and Analysis}
\subsection{Judgment Prediction}
\begin{table}[t]
\centering
\resizebox{\columnwidth}{!}{%
\begin{tabular}{lrrrr}
\hline
\multicolumn{1}{l}{\textbf{Test Data}} &
  \multicolumn{1}{c}{\textbf{\begin{tabular}[c]{@{}c@{}}Macro \\ Precision\end{tabular}}} &
  \multicolumn{1}{c}{\textbf{\begin{tabular}[c]{@{}c@{}}Macro \\ Recall\end{tabular}}} &
  \multicolumn{1}{c}{\textbf{\begin{tabular}[c]{@{}c@{}}Macro \\ F1\end{tabular}}} &
  \multicolumn{1}{c}{\textbf{Accuracy}} \\ \hline
\multicolumn{5}{c}{\textbf{InLegalBert}}                                                                                                                              \\ \hline
ILDC                                                                                          & 0.7209          & 0.7169          & 0.7189          & 0.7172          \\
SCI (2019)                                                                                    & 0.8261          & 0.8255          & 0.8258          & 0.8258          \\
SCI (2020-24)                                                                                 & 0.8515          & 0.8588          & 0.8552          & 0.8720          \\
SCI+HCs (2019)                                                                                & 0.8739          & 0.8735          & 0.8737          & 0.8739          \\
HCs (2020-24)                                                                                 & \textbf{0.8940} & \textbf{0.8943} & \textbf{0.8942} & 0.8945          \\
SCI+HCs+Tribunal (2019)                                                                       & 0.8637          & 0.8634          & 0.8635          & 0.8635          \\
Tribunal (2020-24)                                                                            & 0.8308          & 0.8249          & 0.8278          & 0.8277          \\
\begin{tabular}[c]{@{}l@{}}SCI+HCs+Tribunal+\\ DailyOrders+DistrictCourts (2019)\end{tabular} & 0.8722          & 0.8718          & 0.8720          & 0.8720          \\
Daily\_orders (2020-24)                                                                       & 0.8897          & 0.8869          & 0.8883          & \textbf{0.8955} \\ \hline
\multicolumn{5}{c}{\textbf{InCaseLaw}}                                                                                                                                \\ \hline
ILDC                                                                                          & 0.7347          & 0.7335          & 0.7341          & 0.7337          \\
SCI (2019)                                                                                    & 0.8271          & 0.8272          & 0.8271          & 0.8272          \\
SCI (2020-24)                                                                                 & 0.8449          & 0.8579          & 0.8513          & 0.8670          \\
SCI+HCs (2019)                                                                                & 0.8585          & 0.8570          & 0.8578          & 0.8579          \\
HCs (2020-24)                                                                                 & \textbf{0.8891} & \textbf{0.8898} & \textbf{0.8895} & 0.8898          \\
SCI+HCs+Tribunal (2019)                                                                       & 0.8544          & 0.8521          & 0.8532          & 0.8526          \\
Tribunal (2020-24)                                                                            & 0.8160          & 0.7939          & 0.8048          & 0.8001          \\
\begin{tabular}[c]{@{}l@{}}SCI+HCs+Tribunal+\\ DailyOrders+DistrictCourts (2019)\end{tabular} & 0.8573          & 0.8553          & 0.8563          & 0.8559          \\
Daily\_orders (2020-24)                                                                       & 0.8868          & 0.8795          & 0.8831          & \textbf{0.8911} \\ \hline
\multicolumn{5}{c}{\textbf{XLNet Large}}                                                                                                                              \\ \hline
ILDC                                                                                          & 0.6851          & 0.6850          & 0.6851          & 0.6849          \\
SCI (2019)                                                                                    & 0.8150          & 0.8137          & 0.8143          & 0.8142          \\
SCI (2020-24)                                                                                 & 0.8507          & 0.8562          & 0.8535          & 0.8709          \\
SCI+HCs (2019)                                                                                & 0.8590          & 0.8585          & 0.8588          & 0.8590          \\
HCs (2020-24)                                                                                 & \textbf{0.8848} & 0.8863          & 0.8856          & \textbf{0.8851} \\
SCI+HCs+Tribunal (2019)                                                                       & 0.8580          & 0.8576          & 0.8578          & 0.8577          \\
Tribunal (2020-24)                                                                            & 0.8180          & 0.8053          & 0.8116          & 0.8098          \\
\begin{tabular}[c]{@{}l@{}}SCI+HCs+Tribunal+\\ DailyOrders+DistrictCourts (2019)\end{tabular} & 0.8660          & 0.8655          & 0.8657          & 0.8657          \\
Daily\_orders (2020-24)                                                                       & 0.8832          & \textbf{0.8904} & \textbf{0.8868} & 0.8924  
\\ \hline

\end{tabular}%
}
\caption{Judgment prediction results on the binary task across different court cases and temporal test cases, with models trained on SCI + HCs + Tribunal + Daily Orders and District Court data from \texttt{\named} single split data. The best results are highlighted in bold.}
\label{tab:binary-judgment-results-single}
\end{table}
\begin{table}[t]
\centering
\resizebox{\columnwidth}{!}{%
\begin{tabular}{lrrrr}
\toprule
\multicolumn{1}{l}{\textbf{Models}} &
  \textbf{\begin{tabular}[c]{@{}c@{}}Macro \\ Precision \end{tabular}} &
  \textbf{\begin{tabular}[c]{@{}c@{}}Macro \\ Recall \end{tabular}} &
  \textbf{\begin{tabular}[c]{@{}c@{}}Macro \\ F1 \end{tabular}} &
  \textbf{\begin{tabular}[c]{@{}c@{}}Accuracy \end{tabular}} \\ \midrule

\multicolumn{5}{c}{\textbf{Using LMs on PredEx}}                                                                      \\ \midrule
InLegalBert                                                                  & 75.46 & 75.26 & 75.36 & 75.26 \\ 
InCaseLaw                                                              & 74.21 & 73.95 & 74.08 & 73.95 \\ 
XLNet Large                                                                 & 77.36 & 77.07 & 77.22 & 77.07 \\ 
RoBerta Large                                                           & \textbf{78.31} & \textbf{78.22} & \textbf{78.27} & \textbf{78.22} \\ 
\midrule
\multicolumn{5}{c}{\textbf{Using LLMs on PredEx}}                                                                      \\ \midrule
Zephyr                                                                  & 53.47 & 52.95 & 51.19 & 53.09 \\ 
Gemini pro                                                              & 59.76 & 58.03 & 56.10 & 58.08 \\ 
LLaMa-2                                                                 & 57.32 & 57.23 & 57.13 & 57.26 \\ 
LLaMa-2 SFT                                                             & 51.86 & 51.77 & 51.77 & 51.77 \\ 
\begin{tabular}[c]{@{}l@{}}LLaMa-2 CPT  \\ Prediction only\end{tabular}                                                         & 46.84     & 49.67      & 34.90      & 49.10      \\ 
\begin{tabular}[c]{@{}l@{}}\namel \\ CPT+SFT \\  Prediction only\end{tabular} & 54.39      & 52.04      & 44.62     & 52.04      \\  
\begin{tabular}[c]{@{}l@{}}LLaMa-2 CPT  \\ Prediction + Explanation\end{tabular}                                                         & 45.54     & 49.65      & 35.12      & 50.86      \\ 
\begin{tabular}[c]{@{}l@{}}\namel \\ CPT+SFT \\  Prediction + Explanation \end{tabular} & \textbf{76.23}      & \textbf{76.05 }     & \textbf{76.01 }    & \textbf{76.05}      \\ \midrule
\multicolumn{5}{c}{\textbf{Using LLMs on ILDC\_expert}}                                                                 \\ \midrule
Zephyr                                                                  & 40.11      & 37.04      & 35.24      & 55.56      \\ 
Gemini pro                                                              & 64.67      & 57.41      & 51.39      & 57.41     \\ 
LLaMa-2                                                                 & 31.25 & 42.59 & 32.36 & 42.59 \\ 
LLaMa-2 SFT                                                             & 57.50 & 57.41 & 57.28 & 57.41 \\ 
\begin{tabular}[c]{@{}l@{}}LLaMa-2 CPT  \\ Prediction only\end{tabular}                                                         & 31.25     & 45.45      & 37.04      & 58.82      \\ 
\begin{tabular}[c]{@{}l@{}}\namel \\ CPT+SFT \\  Prediction only\end{tabular} & 59.26      & 59.26      & 59.26     & 59.26      \\ 
\begin{tabular}[c]{@{}l@{}}LLaMa-2 CPT  \\ Prediction + Explanation\end{tabular}                                                         & 26.92     & 50.00      & 35.00      & 53.85      \\ 
\begin{tabular}[c]{@{}l@{}}\namel \\ CPT+SFT \\  Prediction + Explanation \end{tabular} & \textbf{73.01}      & \textbf{72.23 }     & \textbf{71.98 }    & \textbf{72.23}      \\ \bottomrule
\end{tabular}%
}
\caption{Judgment prediction results using different LLMs on PredEx and ILDC datasets. All results are reported in percentages.}
\label{tab:llm-judgment-prediction-results}
\end{table}


\begin{table}[ht]
\centering
\resizebox{\columnwidth}{!}{%
\begin{tabular}{llrrrrrr}
\toprule
\textbf{Models} & \textbf{Metric} & \textbf{Overall} & \textbf{Class 0} & \textbf{Class 1} & \textbf{Class 2} \\ \midrule
\multirow{4}{*}{InLegalBert} 
& Macro Precision & 0.64 & 0.73 & 0.67 & 0.52 \\ 
& Macro Recall & 0.59 & 0.77 & 0.81 & 0.19 \\ 
& Macro F1 & 0.59 & 0.75 & 0.73 & 0.28 \\ \midrule

\multirow{4}{*}{InCaseLaw} 
& Macro Precision & 0.63 & 0.72 & 0.67 & 0.49 \\ 
& Macro Recall & 0.58 & 0.78 & 0.80 & 0.17 \\ 
& Macro F1 & 0.57 & 0.75 & 0.73 & 0.25 \\ \midrule

\multirow{4}{*}{XLNet} 
& Macro Precision & \textbf{0.64} & 0.75 & 0.70 & 0.48 \\ 
& Macro Recall & \textbf{0.61} & 0.80 & 0.82 & 0.22 \\ 
& Macro F1 & \textbf{0.61} & 0.77 & 0.76 & 0.31 \\ \bottomrule

\end{tabular}%
}
\caption{Judgment prediction results on the ternary task on SCI court cases. The best results are in bold.}
\label{tab:judgment-prediction-ternary}
\end{table}
Our experiments, as detailed in Table \ref{tab:binary-judgment-results-single}, which is trained on \texttt{\named} single, and Table \ref{tab:binary-judgment-results-multi} in Appendix, which is trained on \texttt{\named} multi reveal interesting insights into the performance of various models on the Judgment prediction results on the binary task across different court cases and temporal test cases, with models trained on SCI + HCs + Tribunal + Daily Orders and District Court data from \texttt{\named} test data. 

Our findings indicate that contrary to previous research, larger models like XLNet did not consistently outperform smaller models. Instead, models specifically trained on Indian legal data, such as InLegalBERT and InCaseLaw, performed comparably and, in some instances, even surpassed XLNet large. Suggests that the inclusion of domain-specific data significantly enhances performance. The previous best results hovered around 79\% accuracy using XLNet with hierarchical BiGRU, while our best models achieved approximately 90\% accuracy, highlighting a substantial improvement.

While evaluating our model on the ILDC dataset, a noticeable drop in performance was observed compared to its performance on our dataset, \texttt{\named}. A detailed investigation revealed multiple inconsistencies in the ILDC dataset that likely contributed to this issue. These inconsistencies included unremoved results and significant preprocessing errors. For instance, the preprocessing process introduced incorrect replacements for common legal terms, resulting in errors such as "\underline{companycerned}" instead of "concerned", "\underline{companysider}" instead of "consider", and "\underline{numbertion}" instead of "notation". These errors were caused by regex-based transformations in the preprocessing code, such as replacing "co." with "company" and "no." with "number", which inadvertently distorted the original text.


Despite these challenges, as demonstrated in our Ablation study, our model performed robustly when tested on SCI data. The SCI dataset, which is structurally similar to ILDC but more reliable, allowed our model to outperform the results reported in the ~\cite{malik-etal-2021-ildc} paper. This finding highlights the robustness of our model when evaluated on datasets with fewer inconsistencies and better preprocessing. These observations reinforce the importance of dataset quality and preprocessing standards in achieving meaningful advancements in legal judgment prediction tasks.

Table \ref{tab:llm-judgment-prediction-results} showcases the binary prediction results across various LLM-based models. Our findings indicate that while injecting legal knowledge and enhancing reasoning skills on the prediction task did not significantly boost performance, the combination of both on tasks integrating both prediction and explanation yielded more promising results. This improvement is likely attributed to the incorporation of a chain of thought (CoT) process, which aids in making more accurate judgment predictions by facilitating deeper reasoning.

Additionally, we observed that the base LLaMa model struggled in this domain, achieving only 57.26\% accuracy. In comparison, our specialized \texttt{\namel} model, which is a 4-bit quantized version of LLaMa2 adapted through Continual Pretraining (CPT) with domain-specific legal knowledge, significantly outperformed it with an accuracy of 76.05\%. This stark improvement underscores the limitation of the base LLaMa model, which lacks training on Indian-specific legal data, and highlights the effectiveness of our domain-adaptive approach.

Moreover, our comparison with different transformer-based models, including our \texttt{\namel} model, reveals that it is competitive with classifier-based models. This highlights the importance of infusing the model with domain-specific legal knowledge and further refining it through reasoning tasks. 
Table \ref{tab:judgment-prediction-ternary} showcases the results for the ternary judgment prediction task on SCI court cases. Similar trends were observed in the binary prediction tasks, with models trained on the Indian legal corpus showing marked improvements in performance.

\begin{table*}[t]
\centering
\resizebox{\linewidth}{!}{%
\begin{tabular}{lcccccccc}
\toprule
 &
  \multicolumn{5}{c}{\textbf{Lexical Based Evaluation (\%)}} &
  \multicolumn{2}{c}{\textbf{Semantic Evaluation (\%)}} &
  \textbf{Expert Evaluation} \\ \cline{2-9} 
\multirow{-2}{*}{\textbf{Models}} &
  \multicolumn{1}{c}{\textbf{Rouge-1}} &
  \multicolumn{1}{c}{\textbf{Rouge-2}} &
  \multicolumn{1}{c}{\textbf{Rouge-L}} &
  \multicolumn{1}{c}{\textbf{BLEU}} &
  \multicolumn{1}{c}{\textbf{METEOR}} &
  \multicolumn{1}{c}{\textbf{BERTScore}} &
  \multicolumn{1}{c}{\textbf{BLANC}} &
  \textbf{Rating Score} \\ \midrule
 &
  \multicolumn{8}{c}{\textbf{Prediction with Explanation on PredEx \cite{nigam2024legaljudgmentreimaginedpredex}}} \\ \midrule
Gemini pro &
  \multicolumn{1}{c}{30.99} &
  \multicolumn{1}{c}{24.28} &
  \multicolumn{1}{c}{25.93} &
  \multicolumn{1}{c}{8.26} &
  \multicolumn{1}{c}{18.70} &
  \multicolumn{1}{c}{63.29} &
  \multicolumn{1}{c}{17.15} &
  2.24 \\ 
Aalap &
  \multicolumn{1}{c}{27.11} &
  \multicolumn{1}{c}{10.01} &
  \multicolumn{1}{c}{17.03} &
  \multicolumn{1}{c}{3.24} &
  \multicolumn{1}{c}{15.28} &
  \multicolumn{1}{c}{55.41} &
  \multicolumn{1}{c}{7.42} &
  2.46 \\ 
LLaMa-2 &
  \multicolumn{1}{c}{32.11} &
  \multicolumn{1}{c}{18.86} &
  \multicolumn{1}{c}{21.09} &
  \multicolumn{1}{c}{5.99} &
  \multicolumn{1}{c}{17.60} &
  \multicolumn{1}{c}{61.91} &
  \multicolumn{1}{c}{15.07} &
  3.06 \\ 
LLaMa-2 SFT &
  \multicolumn{1}{c}{49.72} &
  \multicolumn{1}{c}{43.21} &
  \multicolumn{1}{c}{\textbf{43.99}} &
  \multicolumn{1}{c}{25.31} &
  \multicolumn{1}{c}{36.30} &
  \multicolumn{1}{c}{\textbf{69.09}} &
  \multicolumn{1}{c}{28.44} &
  2.84 \\ 
LLaMa-2 CPT &
  \multicolumn{1}{c}{33.55} &
  \multicolumn{1}{c}{15.49} &
  \multicolumn{1}{c}{22.87} &
  \multicolumn{1}{c}{8.98} &
  \multicolumn{1}{c}{23.26} &
  \multicolumn{1}{c}{58.34} &
  \multicolumn{1}{c}{11.18} &
  3.26 \\ 
\texttt{InLegalLlama} CPT+SFT &
  \multicolumn{1}{c}{\textbf{50.76}} &
  \multicolumn{1}{c}{\textbf{43.38}} &
  \multicolumn{1}{c}{43.79} &
  \multicolumn{1}{c}{\textbf{25.55}} &
  \multicolumn{1}{c}{\textbf{36.43}} &
  \multicolumn{1}{c}{68.25} &
  \multicolumn{1}{c}{\textbf{29.27}} &
  \textbf{3.54} \\ \midrule
 &
  \multicolumn{8}{c}{\textbf{Prediction with Explanation on ILDC\_expert \cite{vats-etal-2023-llms, malik-etal-2021-ildc}}} \\\midrule 
GPT-3.5 Turbo &
  \multicolumn{1}{c}{\textbf{53.83}} &
  \multicolumn{1}{c}{\textbf{42.67}} &
  \multicolumn{1}{c}{\textbf{45.41}} &
  \multicolumn{1}{c}{28.42} &
  \multicolumn{1}{c}{46.85} &
  \multicolumn{1}{c}{\textbf{72.73}} &
  \multicolumn{1}{c}{33.94} &
  3.60 \\ 
Aalap &
  \multicolumn{1}{c}{29.91} &
  \multicolumn{1}{c}{9.48} &
  \multicolumn{1}{c}{18.08} &
  \multicolumn{1}{c}{4.91} &
  \multicolumn{1}{c}{25.64} &
  \multicolumn{1}{c}{53.79} &
  \multicolumn{1}{c}{9.44} &
  2.30 \\ 
LLaMa-2 &
  \multicolumn{1}{c}{45.26} &
  \multicolumn{1}{c}{24.54} &
  \multicolumn{1}{c}{29.57} &
  \multicolumn{1}{c}{14.85} &
  \multicolumn{1}{c}{34.40} &
  \multicolumn{1}{c}{64.64} &
  \multicolumn{1}{c}{22.12} &
  3.65 \\ 
LLaMa-2 SFT &
  \multicolumn{1}{c}{49.39} &
  \multicolumn{1}{c}{38.05} &
  \multicolumn{1}{c}{39.69} &
  \multicolumn{1}{c}{\textbf{29.18}} &
  \multicolumn{1}{c}{50.75} &
  \multicolumn{1}{c}{68.91} &
  \multicolumn{1}{c}{36.36} &
  3.30 \\ 
LLaMa-2 CPT &
  \multicolumn{1}{c}{30.83} &
  \multicolumn{1}{c}{22.11} &
  \multicolumn{1}{c}{25.50} &
  \multicolumn{1}{c}{14.18} &
  \multicolumn{1}{c}{36.81} &
  \multicolumn{1}{c}{59.29} &
  \multicolumn{1}{c}{25.72} &
  3.41 \\ 
\texttt{InLegalLlama} CPT+SFT &
  \multicolumn{1}{c}{50.88} &
  \multicolumn{1}{c}{40.26} &
  \multicolumn{1}{c}{42.29} &
  \multicolumn{1}{c}{28.20} &
  \multicolumn{1}{c}{\textbf{54.12}} &
  \multicolumn{1}{c}{67.58} &
  \multicolumn{1}{c}{\textbf{40.72}} &
  \textbf{3.67} \\ \toprule
\end{tabular}%
}
\caption{Explanation performance comparison of various models across evaluation metrics. All results are in percentages except for the expert rating scores. The highest scores are bolded.}
\label{tab:explanation-table}
\end{table*}

\subsection{Judgment Prediction with Explanation}
The results, as presented in Table \ref{tab:explanation-table}, offer valuable insights into the comparative performance of machine-generated explanations against those provided by legal experts across various models. These evaluations cover lexical-based, semantic, and expert assessment metrics, specifically using the 50 test cases from the PredEx \cite{nigam2024legaljudgmentreimaginedpredex} and 54 ILDC\_expert \cite{malik-etal-2021-ildc}, comparisons with the \texttt{\namel} model with different settings. This comprehensive evaluation framework allows us to thoroughly assess the models' abilities to generate accurate and contextually relevant explanations.

Additionally, we experimented with the Aalap \cite{tiwari2024aalap} model, which is instruction-tuned on various Indian legal tasks, but it underperforms in this task. This may be due to its lack of focus on generating explanations alongside predictions, a complex requirement that might not have been sufficiently addressed during training. In contrast, comparisons with the \texttt{\namel} model under different settings demonstrate our approach's effectiveness in improving the explainability and accuracy of AI-generated legal judgments.

\subsubsection{Lexical-Based Evaluation}
The performance of LLMs in generating explanations reveals that verbatim matches to reference texts are not consistently high. However, it is important to recognize that these metrics, although useful, do not fully capture the models' capabilities in analyzing legal cases, predicting outcomes, and generating reasoning. Therefore, we also employed Semantic Similarity-Based Evaluation and Expert Score Evaluation to provide a more comprehensive assessment of the models' performance.

\subsubsection{Semantic Evaluation}
The semantic evaluation, particularly utilizing BERTScore, demonstrates better alignment of the generated explanations with the gold standard, indicating a strong semantic understanding of the explanations produced. \texttt{\namel} shows superior performance in terms of semantic similarity. It is important to note that generative models may occasionally produce hallucinated content, which these metrics do not fully capture, emphasizing the need for manual review by legal experts to ensure comprehensive assessment.

\subsubsection{Expert Evaluation}
\label{subsec:expert_evaluation}
Assessing the performance of generative models in the task of LJP requires the insight of domain-specific experts. The expert evaluation, summarized in Table \ref{tab:explanation-table}, indicates that our \texttt{\namel} model, performs exceptionally well, although it sometimes generates truncated or repetitive content. Despite these minor drawbacks, the instruction-tuned model produces fewer non-factual responses and delivers a higher overall quality of explanations compared to other pre-trained models. Notably, models equipped with carefully designed prompts for explanation generation showed improved performance and did not suffer from hallucination issues. The expert ratings, detailed in Table \ref{expert-scores}, further emphasize the effectiveness of our instruction-tuned model, which in some instances even surpasses the quality of explanations provided by human legal experts, achieving an impressive rating score of 4. This highlights the potential of generative models, particularly those enhanced by our approach, in delivering accurate and contextually relevant legal explanations.

%% file: conculsion.tex
\section{Conclusions and Future Work}
In this study, we presented \texttt{\named}, the largest and most diverse dataset of Indian legal cases, alongside \texttt{\namel}, a specialized language model fine-tuned for legal judgment prediction and explanation. Our findings demonstrate that domain-specific models, particularly those enhanced with legal data, significantly outperform generic large language models in both accuracy and quality of explanations provided. Notably, we achieved very good accuracy in the prediction task after including data from all court levels, underscoring the value of comprehensive datasets.

%
Future work will focus on expanding the dataset to include judgments in regional languages, better reflecting India's linguistic diversity. We plan to explore larger and more advanced models, potentially using more efficient quantization techniques and enhanced hardware resources, to better handle complex legal documents. Refining our fine-tuning methodologies by incorporating a broader range of legal documents, such as statutes and contracts, will further enrich the model's knowledge base.

By addressing these challenges and expanding the scope of our research, we aim to enhance the performance and reliability of AI models in the legal domain, contributing to more efficient and accurate legal decision-making processes.


%% file: acknowledgement.tex
\section*{Acknowledgements}

We would like to express our gratitude to the anonymous reviewers for their insightful comments and constructive feedback, which have significantly improved the quality of this work. We also sincerely thank the student research assistants from various law colleges for their invaluable contributions in annotating the documents. Their efforts have been instrumental in the development of this research.  

This work was supported by the ``Research-I Foundation'' at the Dept. of Computer Science and Engineering, IIT Kanpur, which has generously funded the author's conference travel.

%% file: limitation.tex
\section*{Limitations}
Our study faced several significant limitations that influenced both our approach and the findings. One of the primary constraints was using a 4-bit quantized model due to resource constraints, which restricted our ability to leverage larger parametric models, such as those with 70B or 40B parameters. The token limitation and high subscription charges for paid cloud services further exacerbated this issue, limiting our capacity to perform inference and fine-tuning on more advanced models. This limitation likely restricted the full exploration of these models' capabilities, potentially affecting the depth and quality of the insights and performance metrics we could achieve.

Additionally, the resource-intensive nature of obtaining legal expert annotations presented another challenge. The high costs and significant time required for acquiring these annotations made obtaining expert evaluations for the entire PredEx test dataset impractical. As a result, we use the same 50 random documents as used in \cite{nigam2024legaljudgmentreimaginedpredex} for expert review and Likert score evaluations. While necessary, this approach potentially limits the breadth and depth of our expert evaluation, as it does not encompass the entire dataset.

The applicability of LLMs in the legal domain, particularly for tasks involving legal judgment prediction and explanation, remains uncertain based on our findings. While they show proficiency in conversational contexts, their performance in tasks requiring complex logic or specialized knowledge, such as legal reasoning, is less convincing. Analyzing lengthy legal documents and generating predictions and explanations proved challenging for generative models. This challenge is particularly evident when the models must process and understand intricate legal reasoning and contexts.

Lastly, the dataset used in this study comprised only English-language judgments, excluding other regional languages such as Hindi and Bengali. This limitation underscores the need for more inclusive datasets representing the linguistic diversity in legal documents across different jurisdictions.

These limitations highlight the complexities and challenges of applying LLMs to specialized tasks like legal judgment prediction and explanation. They also underscore the need for ongoing research and development to comprehensively enhance AI models' capabilities in interpreting and understanding legal documents and contexts.

%% file: ethics.tex
\section*{Ethics Statement}
In conducting this research, we placed a strong emphasis on ethical considerations, particularly due to the sensitive nature of legal data and the methodologies employed. The \texttt{\named} dataset, used extensively in this study, was sourced from publicly accessible legal search engines, ensuring compliance with data privacy and usage regulations. We have taken steps to remove any meta-information such as judge names, case titles, and case IDs to protect the privacy and confidentiality of the individuals involved.

Furthermore, the computational resources utilized in this study were obtained through ethical and legitimate means. We subscribed to Google Colab Pro and other necessary cloud services, ensuring that all resources used for model training and testing were accessed legally. This financial support not only facilitated our research but also contributed to the sustainability of these services.

In addition to adhering to legal and ethical guidelines in data handling and resource usage, we are committed to transparency and reproducibility in our research. The \texttt{\named} dataset and the code for our models, including \texttt{\namel}, for now, has been made available to promote open science and enable other researchers to replicate and build upon our work.

Finally, we acknowledge the potential societal impact of deploying AI in legal settings. Our models are designed to assist, not replace, human judgment, and we stress the importance of human oversight in any AI-assisted legal decision-making process. We remain committed to ongoing ethical scrutiny as we advance this research field.

%% file: appendix.tex
\section{Ablation Study}
In our ablation study, we investigated the impact of various court-level training data configurations on the performance of our models in the binary classification judgment prediction task. We observed consistent trends across multiple experiments, particularly when analyzing the results from the full \texttt{\named} dataset in relation to subsets that excluded specific court cases. The performance metrics are detailed in Appendix Tables \ref{tab:binary-judgment-results-single-SCI-HCs-Tribunal}, \ref{tab:binary-judgment-results-multi-SCI-HCs-Tribunal}, \ref{tab:binary-judgment-results-single-SCI-HCs}, \ref{tab:binary-judgment-results-multi-SCI-HCs}, \ref{tab:binary-judgment-results-single-SCI}, \ref{tab:binary-judgment-results-multi-SCI}, \ref{tab:binary-judgment-results-ILDC-single}, and \ref{tab:binary-judgment-results-ILDC-multi}. 

Our findings indicate that models trained on High Court cases exhibited the best performance, likely due to the substantial representation of High Court data in the training set. However, when these models were evaluated on the ILDC dataset, a noticeable drop in performance was observed. This suggests that while the models excel in familiar contexts, they struggle to generalize to datasets with different characteristics or distributions.

We also conducted experiments on the ternary judgment prediction task, incorporating additional court cases, with results presented in Appendix Tables \ref{tab:judgment-prediction-ternary-sci-HCs}, \ref{tab:judgment-prediction-ternary-sci-hcs-tribunals}, and \ref{tab:judgment-prediction-ternary-sci-hcs-tribunals-dailyorders}. These experiments further reinforced the significance of including diverse court data and highlighted the benefits of utilizing large volumes of training data. 

Our findings emphasize the importance of a diverse and comprehensive dataset. Including a broad spectrum of court cases not only enhances the model's understanding of various judicial contexts but also significantly improves its performance metrics. In particular, the models trained on the most extensive datasets achieved an F1 score of around 90\%, underscoring the critical role of data diversity and volume in achieving high accuracy.

Overall, the ablation study illustrates that the diversity and volume of training data play a crucial role in enhancing model performance, particularly in the context of legal judgment prediction tasks. Future work should continue to explore the impact of various data configurations to further optimize model accuracy and generalization capabilities.

\section{Rating Score Description}
\label{section:rating-score-description}
The evaluation of the explanations generated by the models was conducted using a 1–5 Likert scale, where each score reflects the quality and relevance of the provided explanation. The criteria for rating are as follows:

\textbf{[1]:} The explanation is entirely incorrect or fails to provide any relevant information. This score indicates that the response does not address the case judgment in any meaningful way.

\textbf{[2]:} The response is irrelevant or demonstrates a misunderstanding of the case judgment. A rating of 2 suggests that while some effort was made to respond, the explanation does not accurately reflect the case details.

\textbf{[3]:} The explanation is partially accurate but lacks critical details. This score indicates that the response contains some correct information, but it is insufficient for a complete understanding of the case judgment.

\textbf{[4]:} The response is generally accurate and relevant, comparable to the ground truth. A rating of 4 signifies that the explanation aligns well with the expected outcomes and provides a solid understanding of the case.

\textbf{[5]:} The explanation is fully accurate, relevant, and potentially superior to the expert's explanation. This highest rating reflects an exceptional response that not only meets the criteria of accuracy and relevance but also offers insights that exceed standard expert evaluations.

\section{Experimental Setup and Hyper-parameters}
\label{sec:Experimental-setup}

\subsection{Transformers Training Hyper-parameters}
For model training, we used a batch size of 16, the Adam optimizer \cite{kingma2014adam}, and a learning rate of 2e-6. The training was conducted over 3 epochs on the \texttt{\named} train dataset. The remaining hyperparameters were set to their default values as provided by the HuggingFace library.

\subsection{\namel Training Procedure}
The fine-tuning of the \texttt{\namel} model was conducted using the LLaMa 2 7B model architecture, with the model loaded in Bfloat16 precision. The training was done in Google Colab Pro, utilizing a single A100 GPU with 40GB of memory. Given the constraints of limited computational resources, we carefully selected parameters to fully utilize the available compute power. This setup enabled us to develop a highly capable model within a reasonable time frame of 48 hours, incurring a cost of approximately \$59. During the training process, the maximum token length was set at 2,096. We employed the Low-Rank Adaptation (LoRA) technique, initializing the LoRA rank at 16 and setting the alpha parameter to 64, with a dropout rate of 0.1. This configuration was applied to all layers of the model, aiming to achieve performance comparable to a fully fine-tuned model. The integration of flash-attention 2 significantly improved the training speed.

The optimization process utilized a Paged Adam 32-bit optimizer with a learning rate of 1e-4, alongside a ``cosine" learning rate scheduler. The gradient accumulation steps were set to 4, and the warm-up ratio was established at 0.05. We employed DeepSpeed Stage 3 optimization, with a per-device batch size of 4. The model was trained for a total of 3,000 steps, which corresponds to approximately 0.347 epochs.


\section{Hallucination}
\label{sec:hallucination}
In our study, we address the issue of hallucinations in model-generated text, which is a prevalent challenge when using large language models for generating legal judgments. Hallucinations occur when the model produces information that is false or irrelevant, not supported by the input data. A sample of hallucination has been provided in Appendix Table \ref{cpt-hallucination-example}. To tackle this issue, we employed a specialized fine-tuning strategy aimed at significantly reducing such errors. A detailed comparative analysis provided in Appendix \ref{subsec:CPT-hallucination} 
 - CPT LLaMa-2 hallucinations, which 
 highlights the effectiveness of these strategies. This analysis illustrates how fine-tuning and instruction-tuning, specifically tailored to the legal domain, can help minimize hallucinations, resulting in outputs that are clearer, more accurate, and legally coherent.


\subsection{CPT LLaMa-2 hallucinations}
\label{subsec:CPT-hallucination}
In the subsection, we conduct a thorough comparison between ground truth and fine-tuned models to demonstrate some samples where the CPT model showed signs of hallucination. Table \ref{cpt-hallucination-example} in the Appendix presents an extensive analysis of the model, illustrating the performance of the LLaMa-2 pre-trained model on legal judgment prediction with explanation tasks in our PredEx dataset. In the given table, this can be observed that the model did not only produce incorrect judgment but also delivered the wrong explanation, if not repeating the sentences and printing a random repetitive set of numbers. Some examples in the table show the model hallucinating by repeating a statement irrelevant to the case information. This comparison specifically focuses on how instruction-tuning can mitigate the issues of inaccuracies often found in the responses generated by pre-trained Large Language Models (LLMs).

\subsection{CPT LLaMa-2 vs \namel (CPT+SFT)}
\label{subsec:CPT-vs-SFT}
In the subsection, we conduct a detailed comparison between LLaMa-2-7B CPT and \namel (CPT+SFT) to demonstrate some samples where the fine-tuned model \namel performed better than LLaMa-2 CPT in all aspects including prediction and explanation. Table \ref{tab:ca_analysis} in the Appendix presents an extensive analysis of the model, illustrating the performance of the LLaMa-2 pre-trained model against the \namel model on legal judgment prediction with explanation tasks in our PredEx dataset. In the given table, it can be observed that the performance of the LLaMa-2 CPT looks good at once but when we refer to the explanation, we realize that the explanation given for the prediction does not follow the context of the legal judgment. On the other hand, \namel maintains the contextual information and retains a lot of information from the legal judgment.As a result, \namel outperforms the LLaMa-2 CPT with better context information and reasonable facts inclusion.


\begin{table*}[ht]
\centering
\tiny
\resizebox{\textwidth}{!}{%
\begin{tabular}{|c|l|}
\hline

\textbf{Model} &
  \multicolumn{1}{c|}{\textbf{Decision and Explanation}} \\ \hline
 
Ground Truth &
  { \begin{tabular}[c]{@{}l@{}}Decision: Rejected. \\Explanation:
  \\IT first requires to be noticed that the above two agreements\\ dated November 1, 1966 and may 13, 1967, are agreements made otherwise \\than in the course of conciliation proceedings and are of the nature referred\\ to in(1) of S. 18 Admittedly, these agreements were not signed on behalf of\\ the workmen who were were not members of the 2nd respondent Union.\\  .........................\\ As we are unable to accept that contention, the third contention alsowe\\ have held that these agreements have ceased to be in operation and did \\not require to be terminated, this contention must fail. We have\\ already recorded reasons in that connection in the previous discussion.\end{tabular}} \\ \hline
Llama2-7B CPT &
  { \begin{tabular}[c]{@{}l@{}}Decision: No clear decision generated.\\ Explanation:\\ 1.0 The order is upheld. 1.1 The order is upheld. 1.2 The order is upheld. \\1.3 The order is upheld. 1.4 The order is upheld. 1.5 The order is upheld.\\ 1.6 The order is upheld. 1.7 The order is upheld. 1.8 The order is upheld. \\1.9 The order is upheld. 1.10 The order is upheld. ............. \\1.73 The order is upheld. 1.74 The order is upheld. 1.75 The order is upheld. \\1.76 The order is upheld. 1.77 The order is upheld. 1.78 The order is upheld.\\ 1.79 The order is upheld. 1.80 The order is upheld. 1.81 The order is upheld.\\ 1.82 The order is upheld. 1.83 The order is upheld. 1.\end{tabular}} \\ \hline

Ground Truth &
  { \begin{tabular}[c]{@{}l@{}}Decision: Rejected \\ Explanation: 
  \\For proper appreciation of the contentions advanced by the parties\\ before us, it is necessary to refer to some of the provisions of the Act.\\ Section 160(1)(i) of the Act prov ides that in respect of the income of\\ a non-resident specified in sub-section (1) of section 9 of the Act, the\\ agent of the non-resident, including a person who is treated as\\ an agent under section 163 is a representative assessee\\...................\\The above cable shows that the London Solicitors had sought information about\\ the suits in Calcutta t o enable them to engage Mr. Blanco White to plead\\ on behalf of the German Corporation in the said suits. On December 23, 1969,\\ the London Solicitors wrote a letter to the appellants in which it\\ was stated that the copies of certain documents sent by the appellants had\\ been handed over to Mr. Blanco White in addition to copies of certain\\ other documents which they themselves had handed over to him.\end{tabular}} \\ \hline
Llama2-7B CPT &
  {\begin{tabular}[c]{@{}l@{}}Decision: Rejected \\Explanation:
  \\The appellants are a partnership firm and the partners are Mr. and\\ Mr. D.K.Jain. Mr. D.K.Jain is a non-resident Indian.Income from the\\ partnership firm is subject to tax in India under the provisions of the\\ Income-tax Act, 1961....................The firm has no in Germany.The firm has no\\ in Germany.The firm has no in Germany.The firm has\\ no in Germany.The firm has no in Germany.The firm has no in Germany.The\\ firm has no in Germany.The firm has no in Germany.The firm has no in\\ Germany.The firm has no in Germany.The firm has no in Germany.The\end{tabular}} \\ \hline
\end{tabular}%
}
\caption{We observe 'reject' judgments in the response generated by LLaMa-7B CPT with varying explanations, conclusions, and repeating responses.}
\label{cpt-hallucination-example}
\end{table*}

\begin{table*}[t]
\centering
\resizebox{\textwidth}{!}{%
\begin{tabular}{|c|l|l|ll}
\cline{1-3}
\textbf{Model}                                                                          & \multicolumn{1}{c|}{\textbf{Decision and Explanation}}                  & \multicolumn{1}{c|}{\textbf{Explanation of error}}                      &  &  \\ \cline{1-3}
Ground Truth                                                                                & {\begin{tabular}[c]{@{}l@{}}Case decision: Accepted\\ Explanation: \\ In our opinion, the High Court should have examined these\\ questions and especially the question as to whether the had been able to\\ substantiate the claim that it was in possession of the entire underground \\rights in respect of Ac.covered by the agreement of 1934. The agreement\\ of 1934 bound the Government and the Hindu undivided family: the was \\not a party to it. Even if the Hindu undivided family continued in\\ possession of some of the mines, seams and veins after October 1, 1947,\\ it is a moot question whether such possession meant possession of the\\ seams, mines, and veins which were not actually being worked. A further\\ question would arise as to how and what right the company could lawfully\\ claim in respect of the colliery by virtue of the deed of assignment in 1950.\\ Without an adjudication of these questions the High Court was not justified\\ in directing the State and other authorities to put the company in possession\\ of the area described as a virgin area. Whether the case is a fit for\\ examination on affidavits will be for the High Court to consider.\end{tabular}} & &  &  \\ \cline{1-3}
Llama-2-7B CPT                                                                                  & { \begin{tabular}[c]{@{}l@{}}Case decision: Accepted\\ Explanation: 
\\1.The petitioner was in actual physical of the area which was\\ taken over by the State Government on February 28, 1970. The High\\ Court was not justified in directing the State and other authorities to\\ put the petitioner in the physical area. The State Government had\\ taken over the area for the purpose of the public interest. The High Court\\ should have refrained from granting any relief to the petitioner.
\\2. The respondent company was not in actual physical of the area which\\ was taken over by the State Government on February 28, 1970. The High\\ Court was not justified in directing the State and other authorities to put\\ the petitioner in physical of the area. The State Government had taken\\ over the area for the purpose of the public interest. The High Court\\ should have refrained from granting any relief to the petitioner.
\\
...
\\12. The High Court should have examined the questions of whether the\\ petitioner was in actual physical of the area which was taken over by\\ the State Government on February 28, 1970, and whether the State\\ Government had taken over the area for the purpose of the public\\ interest. The High Court should have refrained from granting any\\ relief to the petitioner.
\\13 The High Court should have examined the questions whether the\\ petitioner was in actual physical of the area which was taken\end{tabular}}                                                     

& \begin{tabular}[c]{@{}l@{}} The vanilla incorrectly explains the\\ explanation of the case text. The reason\\ behind hallucination is lack of contextual\\ understanding of the case information,\\ overemphasis on certain parts of the\\ input, and missing information such as\\ the agreement of 1934 and 1950. One of the\\ reasons could also be the complex legal\\ language that the model could not fully grasp.
\end{tabular} &  &  \\ \cline{1-3}

\begin{tabular}[c]{@{}c@{}}\namel \\ (CPT+SFT)\\ Instruction-tuning\\ for prediction \\ with explanation\\ task\end{tabular} & \begin{tabular}[c]{@{}l@{}}Case decision: Accepted\\ Explanation: \\ Further there was no agreement as regards the royalties\\ to be paid under the renewed lease.
\\23. With regard to the right claimed under the deed of assignment\\ it was argued that the assignment could take place only when the\\ lease was in force and as the lease of 1917 had expired in 1947\\ no valid assignment could be taken place if indeed, there was, in fact, \\ any assignment of the colliery as alleged.
\\24. It was urged that without examining those questions and without\\ coming to the conclusion that the petitioner-company had by their\\ petition disclosed a legal right in them with regard to the colliery\\ or to continue in possession thereof the High Court was not competent\\ to grant them any relief as it had purported to do. It was submitted\\ that at the most there was a term in the agreement of 1934 read with\\ the lease of 1917 which entitled the Hindu undivided family to sue\\ for specific performance of the contract to renew for a further term.\\ The demise in its favor of the veins etc., under the lease of 1917 had\\ come to an end and even if the person in possession of the seams\\ and veins which were being worked could not be dispossessed by an\\ order of the nature disclosed in the letter of February 1970, the \\petitioner-company had not been able to show any right in it to claim\\ possession of the virgin area, that it to say, the seems, veins, mines\\, etc., of which they were not in physical possession and which they\\ were not working at the moment, when the impugned order was served on\\ them.
\\25. In our opinion, the High Court should have examined these questions\\ and especially the question as to whether the petitioner-company had been\\ able to substantiate the claim that it was in possession of the entire\\ underground rights in respect of Ac. 871-49 covered by the agreement\\ of 1934. The agreement of 1934 bound the Government and the Hindu\\ undivided family: the petitioner company was not a party to it. Even if\\ the Hindu undivided family continued in possession of some of the mines\end{tabular}                                      & \begin{tabular}[c]{@{}l@{}}The Finetuned model correctly explains\\ the reasoning behind the case judgment \\ comparatively better. The reasoning contains\\ better context information and is organized\\ logically. This explanation includes specific legal\\ arguments related to the agreements.\\ The explanation also compares the petitioner's\\ claim with actual agreements and classes. \end{tabular}&  &  \\ \cline{1-3}
\end{tabular}%
}
\caption{Comparative analysis of responses generated by LLaMa-2-7B (CPT) and \namel (CPT+SFT).}
\label{tab:ca_analysis}
\end{table*}


\begin{table*}[t]
\centering
\resizebox{\textwidth}{!}{%
\begin{tabular}{|
>{}l |}
\hline
\textbf{CASE NO:} \\ \hline
 CIVIL APPEAL NOS. 3088-3089 OF 2020\\ \hline
\textbf{APPELLANTS:} \\ \hline
BHARTI AIRTEL LIMITED AND ANOTHER  \\ \hline
\textbf{RESPONDENT:} \\ \hline
VIJAYKUMAR V. IYER AND OTHERS \\ \hline
\textbf{DATE OF JUDGMENT:} \\ \hline
03/01/2024 \\ \hline
{ \textbf{BENCH:}} \\ \hline
{ Dipankar Datta} \\ \hline
\textbf{CASE TEXT:} \\ \hline

{ \begin{tabular}[c]{@{}l@{}}
... The present appeals raise an interesting question on the right to claim set-off in the Corporate Insolvency \\Resolution Process when the Resolution Professional proceeds in terms of clause (a) to sub-section (2) of\\ \textcolor{blue}{Section 25 of the Insolvency and Bankruptcy Code, 20161} to take custody and control of all the assets of \\the corporate debtor .... \\ \\ 

...The NCLAT vide order dated 17th May 2019 allowed the appeal, inter alia, holding that set-off is violative of the \\basic principles and protection accorded under any insolvency law. Set-off is antithetical to the objective of the IBC. \\Reference was made to the non-obstante provisions in the form of \textcolor{blue}{Section 238 of the IBC}. As moratorium under\\ \textcolor{blue}{Section 14(4)} applies till the date of completion of the Corporate Insolvency Resolution Process, which is till the\\ resolution plan...  \\ \\ 

...The judgment of this Court in The \textcolor{magenta}{Official Liquidator of High Court of Karnataka v. Smt. V. Lakshmikutty53} \\ had applied \textcolor{blue}{Section 46 of the Provincial Insolvency Act, 1920} and had accordingly permitted insolvency \\set-off on interpretation and application of \textcolor{blue}{Sections 529 and 530 of the Companies Act, 1956}. In that context, \\it is observed that the English courts, on \textcolor{magenta}{52 Career Institute Educational Society v. Om Shree Thakurji Educational}\\ \textcolor{magenta}{Society, 2023} SCC OnLine SC 586....\\ \\

... Thereupon the Airtel entities made a payment of Rs.341.80 crores due to the Aircel entities on 10th January 2019.\\ The balance amount of Rs.145.20 crores was set-off by the Airtel entities on the ground that this amount was owned\\ by the Aircel entities to the Airtel entities. According to Airtel entities, Rs.145.20 crores was the adjusted or the net \\amount payable by the Aircel entities towards operational charges, SMS charges and interconnect usage charges to\\ the Airtel entities ...\end{tabular}} \\ \hline
\textbf{JUDGEMENT:} \\ \hline
{ \begin{tabular}[c]{@{}l@{}}.... Having considered the contentions raised by the appellant Airtel entities in detail, and in light of the provisions\\ of the IBC relating to the Corporate Insolvency Resolution Process, \textcolor{red}{we do not find any merit in the present appeals}\\ \textcolor{red}{and the same are dismissed}. There will be no order as to costs....\end{tabular}} \\ \hline
\end{tabular}%
}
\caption{Example of Indian Case Structure. Sections referenced are highlighted in blue, previous judgments cited are in magenta, and the final decision is indicated in red.}
\label{case-example}
\end{table*}

\begin{table*}[ht]
    \centering
    \begin{tabular}{|p{0.8\textwidth}|}
    \hline
{\bf Template 1 (prediction + explanation)}\\
\hline
   
{\bf prompt} = f``````Task: Given a Supreme Court of India case proceeding enclosed in angle brackets $<$ $>$, your task is to predict the decision of the case (with respect to the appelant) and provide an explaination for the decision.\\

{\bf Prediction}: Given a case proceeding, the task is to predict the decision 0 or 1, where the label 1 corresponds to the acceptance of the appeal/petition of the appellant/petitioner and the label 0 corresponds to the rejection of the appeal/petition of the appellant/petitioner,  Explanation: The task is to explain how you arrived at the decision by predicting important sentences that lead to the decision. \\

{\bf Context}: Answer in a consistent style as shown in the following two examples: \\

  {\bf case\_proceeding}: \# case\_proceeding example 1\\

  {\bf Prediction}: \# example 1 prediction \\

  {\bf Explanation}: \# example 1 explanation\\

  {\bf case\_proceeding}: \# case\_proceeding example 2\\

 {\bf  Prediction}: \# example 2 prediction\\

 {\bf Explanation}: \# example 2 explanation\\

{\bf Instructions}: Learn from the above two examples and perform the task for the following case proceeding. \\

case\_proceeding: $<$\{case\_proceeding\}$>$\\

Format your output in list format: [prediction, explanation]''''''\\

\hline
    {\bf Template 2 (prediction only)}\\
    \hline
   
{\bf prompt} = f``````Task: Given a Supreme Court of India case proceeding enclosed in angle brackets $<$ $>$, your task is to predict the decision of the case (with respect to the appellant).\\

{\bf Prediction}: Given a case proceeding, the task is to predict the decision 0 or 1, where the label 1 corresponds to the acceptance of the appeal/petition of the appellant/petitioner and the label 0 corresponds to the rejection of the appeal/petition of the appellant/petitioner \\

{\bf Context}: Answer in a consistent style as shown in the following two examples: \\

  {\bf case\_proceeding}: \# case\_proceeding example 1\\

  {\bf Prediction}: \# example 1 prediction \\

  {\bf case\_proceeding}: \# case\_proceeding example 2\\

  {\bf Prediction}: \# example 2 prediction\\

{\bf Instructions}: Learn from the above two examples and perform the task for the following case proceeding. \\

{\bf case\_proceeding}: $<$\{case\_proceeding\}$>$\\

Give the output predicted case decision as either 0 or 1.''''''\\

\hline

    \end{tabular}
    \caption{Prompts for Judgment Prediction taken from \cite{vats-etal-2023-llms}.}
    \label{tab:judgment_prediction_prompts_few}
\end{table*}

\begin{table*}[ht]
    \centering
    \begin{tabular}{|p{0.8\textwidth}|}
    \hline
    {\bf Template 3 (prediction only)}\\
    \hline
   
{\bf prompt} = f``````
\#\#\# {\bf Instructions}: Analyze the case proceeding and predict whether the appeal/petition will be rejected (0) or accepted (1). \\

\#\#\# \textbf{Input}: $<$\{case\_proceeding\}$>$\\

\#\#\# Response:
  ''''''\\
\hline
    {\bf Template 4 (prediction with explanation)}\\
    \hline
   
{\bf prompt} = f``````
\#\#\# {\bf Instructions}: Analyze the case proceeding and predict whether the appeal/petition will be accepted (1) or rejected (0), and subsequently provide an explanation behind this prediction with important textual evidence from the case. \\

\#\#\# \textbf{Input}: $<$\{case\_proceeding\}$>$\\

\#\#\# Response:
  ''''''\\
\hline

    \end{tabular}
    \caption{Prompts for Judgment Prediction used for instruction fine-tuned models taken from \cite{nigam2024legaljudgmentreimaginedpredex}. Instructions were randomly chosen from Table \ref{Instruction-sets}.}
    \label{tab:judgment_prediction_prompts_zero}
\end{table*}

\begin{table*}[ht]
\centering
\resizebox{0.95\textwidth}{!}{%
\tiny
\begin{tabular}{|cl|}
\hline
\multicolumn{2}{|c|}{\textbf{\textcolor{blue}{Instruction sets for Predicting the Decision}}} \\ \hline
\multicolumn{1}{|c|}{1} &
  Analyze the case proceeding and predict whether the appeal/petition will be accepted (1) or rejected (0). \\ \hline
\multicolumn{1}{|c|}{2} &
  \begin{tabular}[c]{@{}l@{}}Based on the information in the case proceeding, determine the likely outcome: acceptance (1) or \\ rejection (0) of the appellant/petitioner's case.\end{tabular} \\ \hline
\multicolumn{1}{|c|}{3} &
  Review the case details and predict the decision: will the court accept (1) or deny (0) the appeal/petition? \\ \hline
\multicolumn{1}{|c|}{4} &
  \begin{tabular}[c]{@{}l@{}}Considering the arguments and evidence in the case proceeding, predict the verdict: is it more likely to be in \\ favor (1) or against (0) the appellant?\end{tabular} \\ \hline
\multicolumn{1}{|c|}{5} &
  \begin{tabular}[c]{@{}l@{}}Examine the details of the case proceeding and forecast if the appeal/petition stands a chance of being \\ upheld (1) or dismissed (0).\end{tabular} \\ \hline
\multicolumn{1}{|c|}{6} &
  \begin{tabular}[c]{@{}l@{}}Assess the case proceedings and provide a prediction: is the court likely to rule in favor of (1) or against (0)\\ the appellant/petitioner?\end{tabular} \\ \hline
\multicolumn{1}{|c|}{7} &
  \begin{tabular}[c]{@{}l@{}}Interpret the case information and speculate on the court's decision: acceptance (1) or rejection (0) of the \\ presented appeal.\end{tabular} \\ \hline
\multicolumn{1}{|c|}{8} &
  \begin{tabular}[c]{@{}l@{}}Given the specifics of the case proceeding, anticipate the court's ruling: will it favor (1) or oppose (0) the \\ appellant’s request?\end{tabular} \\ \hline
\multicolumn{1}{|c|}{9} &
  \begin{tabular}[c]{@{}l@{}}Scrutinize the evidence and arguments in the case proceeding to predict the court's decision: will the appeal\\  be granted (1) or denied (0)?\end{tabular} \\ \hline
\multicolumn{1}{|c|}{10} &
  \begin{tabular}[c]{@{}l@{}}Analyze the legal arguments presented and estimate the likelihood of the court accepting (1) or rejecting (0) \\ the petition.\end{tabular} \\ \hline
\multicolumn{1}{|c|}{11} &
  \begin{tabular}[c]{@{}l@{}}From the information provided in the case proceeding, infer whether the court's decision will be positive (1) \\ or negative (0) for the appellant.\end{tabular} \\ \hline
\multicolumn{1}{|c|}{12} &
  \begin{tabular}[c]{@{}l@{}}Evaluate the arguments and evidence in the case and predict the verdict: is an acceptance (1) or rejection\\ (0) of the appeal more probable?\end{tabular} \\ \hline
\multicolumn{1}{|c|}{13} &
  \begin{tabular}[c]{@{}l@{}}Delve into the case proceeding and predict the outcome: is the judgment expected to be in support (1) or \\ in denial (0) of the appeal?\end{tabular} \\ \hline
\multicolumn{1}{|c|}{14} &
  \begin{tabular}[c]{@{}l@{}}Using the case data, forecast whether the court is likely to side with (1) or against (0) the \\ appellant/petitioner.\end{tabular} \\ \hline
\multicolumn{1}{|c|}{15} &
  \begin{tabular}[c]{@{}l@{}}Examine the case narrative and anticipate the court's decision: will it result in an approval (1) or \\ disapproval (0) of the appeal?\end{tabular} \\ \hline
\multicolumn{1}{|c|}{16} &
  \begin{tabular}[c]{@{}l@{}}Based on the legal narrative and evidentiary details in the case proceeding, predict the court's stance: \\ favorable (1) or unfavorable (0) to the appellant.\end{tabular} \\ \hline
\multicolumn{2}{|c|}{\textbf{\textcolor{blue}{Instruction sets for Integrated Approach for Prediction and Explanation}}} \\ \hline
\multicolumn{1}{|c|}{1} &
  \begin{tabular}[c]{@{}l@{}}First, predict whether the appeal in case proceeding will be accepted (1) or not (0), and then explain the \\ decision by identifying crucial sentences from the document.\end{tabular} \\ \hline
\multicolumn{1}{|c|}{2} &
  \begin{tabular}[c]{@{}l@{}}Determine the likely decision of the case (acceptance (1) or rejection (0)) and follow up with an \\ explanation highlighting key sentences that support this prediction.\end{tabular} \\ \hline
\multicolumn{1}{|c|}{3} &
  \begin{tabular}[c]{@{}l@{}}Predict the outcome of the case proceeding (1 for acceptance, 0 for rejection) and subsequently provide an\\  explanation based on significant sentences in the proceeding.\end{tabular} \\ \hline
\multicolumn{1}{|c|}{4} &
  \begin{tabular}[c]{@{}l@{}}Evaluate the case proceeding to forecast the court's decision (1 for yes, 0 for no), and elucidate the \\ reasoning behind this prediction with important textual evidence from the case.\end{tabular} \\ \hline
\multicolumn{1}{|c|}{5} &
  \begin{tabular}[c]{@{}l@{}}Ascertain if the court will uphold (1) or dismiss (0) the appeal in the case proceeding, and then clarify \\ this prediction by discussing critical sentences from the text.\end{tabular} \\ \hline
\multicolumn{1}{|c|}{6} &
  \begin{tabular}[c]{@{}l@{}}Judge the probable resolution of the case (approval (1) or disapproval (0)), and elaborate on this forecast\\  by extracting and interpreting significant sentences from the proceeding.\end{tabular} \\ \hline
\multicolumn{1}{|c|}{7} &
  \begin{tabular}[c]{@{}l@{}}Forecast the likely verdict of the case (granting (1) or denying (0) the appeal) and then rationalize your \\ prediction by pinpointing and explaining pivotal sentences in the case document.\end{tabular} \\ \hline
\multicolumn{1}{|c|}{8} &
  \begin{tabular}[c]{@{}l@{}}Assess the case to predict the court's ruling (favorably (1) or unfavorably (0)), and then expound on \\ this prediction by highlighting and analyzing key textual elements from the proceeding.\end{tabular} \\ \hline
\multicolumn{1}{|c|}{9} &
  \begin{tabular}[c]{@{}l@{}}Decide if the appeal in the case proceeding is more likely to be successful (1) or unsuccessful (0), and \\ then justify your decision by focusing on essential sentences in the document.\end{tabular} \\ \hline
\multicolumn{1}{|c|}{10} &
  \begin{tabular}[c]{@{}l@{}}Conjecture the end result of the case (acceptance (1) or non-acceptance (0) of the appeal), followed by \\ a detailed explanation using crucial sentences from the case proceeding.\end{tabular} \\ \hline
\multicolumn{1}{|c|}{11} &
  \begin{tabular}[c]{@{}l@{}}Predict whether the case will result in an affirmative (1) or negative (0) decision for the appeal, and then \\ provide a thorough explanation using key sentences to support your prediction.\end{tabular} \\ \hline
\multicolumn{1}{|c|}{12} &
  \begin{tabular}[c]{@{}l@{}}Estimate the outcome of the case (positive (1) or negative (0) for the appellant) and then give a reasoned \\ explanation by examining important sentences within the case documentation.\end{tabular} \\ \hline
\multicolumn{1}{|c|}{13} &
  \begin{tabular}[c]{@{}l@{}}Project the court's decision (favor (1) or against (0) the appeal) based on the case proceeding, and \\ subsequently give an in-depth explanation by analyzing relevant sentences from the document.\end{tabular} \\ \hline
\multicolumn{1}{|c|}{14} &
  \begin{tabular}[c]{@{}l@{}}Make a prediction on the court's ruling (acceptance (1) or rejection (0) of the petition), and then dissect \\ the proceeding to provide a detailed explanation using key textual passages.\end{tabular} \\ \hline
\multicolumn{1}{|c|}{15} &
  \begin{tabular}[c]{@{}l@{}}Speculate on the likely judgment (yes (1) or no (0) to the appeal) and then delve into the case proceeding \\ to elucidate your prediction, focusing on critical sentences.\end{tabular} \\ \hline
\multicolumn{1}{|c|}{16} &
  \begin{tabular}[c]{@{}l@{}}Hypothesize the court's verdict (affirmation (1) or negation (0) of the appeal), and then clarify this \\ hypothesis by interpreting significant sentences from the case proceeding.\end{tabular} \\ \hline
\end{tabular}%
}
 \caption{Instruction Sets for Case Prediction and Explanations taken from \cite{nigam2024legaljudgmentreimaginedpredex}.}
\label{Instruction-sets}
\end{table*}
\begin{table}[t]
\centering
\resizebox{\columnwidth}{!}{%
\begin{tabular}{lrrrr}
\hline
\multicolumn{1}{l}{\textbf{Test Data}} &
  \multicolumn{1}{r}{\textbf{\begin{tabular}[c]{@{}c@{}}Macro \\ Precision\end{tabular}}} &
  \multicolumn{1}{r}{\textbf{\begin{tabular}[c]{@{}c@{}}Macro \\ Recall\end{tabular}}} &
  \multicolumn{1}{r}{\textbf{\begin{tabular}[c]{@{}c@{}}Macro \\ F1\end{tabular}}} &
  \multicolumn{1}{r}{\textbf{Accuracy}} \\ \hline
\multicolumn{5}{c}{\textbf{InLegalBert}}                                                                                                                              \\ \hline
ILDC                                                                                          & 0.7486          & 0.7464          & 0.7475          & 0.7462          \\
SCI (2019)                                                                                    & 0.8391          & 0.8391          & 0.8391          & 0.8388          \\
SCI (2020-24)                                                                                 & 0.8712          & 0.8875          & 0.8793          & 0.8913          \\
SCI+HCs (2019)                                                                                & 0.8843          & 0.8831          & 0.8837          & 0.8837          \\
HCs (2020-24)                                                                                 & \textbf{0.9006} & \textbf{0.9018} & \textbf{0.9012} & \textbf{0.9012} \\
SCI+HCs+Tribunal (2019)                                                                       & 0.8753          & 0.8749          & 0.8751          & 0.8751          \\
Tribunal (2020-24)                                                                            & 0.8545          & 0.8527          & 0.8536          & 0.8541          \\
\begin{tabular}[c]{@{}l@{}}SCI+HCs+Tribunal+\\ DailyOrders+DistrictCourts (2019)\end{tabular} & 0.8800          & 0.8795          & 0.8797          & 0.8797          \\
Daily\_orders (2020-24)                                                                       & 0.8734          & 0.8801          & 0.8768          & 0.8830          \\ \hline
\multicolumn{5}{c}{\textbf{InCaseLaw}}                                                                                                                                \\ \hline
ILDC                                                                                          & 0.7090          & 0.7085          & 0.7088          & 0.7086          \\
SCI (2019)                                                                                    & 0.8209          & 0.8209          & 0.8209          & 0.8209          \\
SCI (2020-24)                                                                                 & 0.8453          & 0.8659          & 0.8555          & 0.8681          \\
SCI+HCs (2019)                                                                                & 0.8564          & 0.8543          & 0.8554          & 0.8554          \\
HCs (2020-24)                                                                                 & \textbf{0.8708} & \textbf{0.8717} & \textbf{0.8712} & \textbf{0.8690} \\
SCI+HCs+Tribunal (2019)                                                                       & 0.8603          & 0.8593          & 0.8598          & 0.8596          \\
Tribunal (2020-24)                                                                            & 0.8451          & 0.8412          & 0.8431          & 0.8433          \\
\begin{tabular}[c]{@{}l@{}}SCI+HCs+Tribunal+\\ DailyOrders+DistrictCourts (2019)\end{tabular} & 0.8578          & 0.8565          & 0.8571          & 0.8570          \\
Daily\_orders (2020-24)                                                                       & 0.8457          & 0.8609          & 0.8532          & 0.8564          \\ \hline
\multicolumn{5}{c}{\textbf{XLNet Large}}                                                                                                                              \\ \hline
ILDC                                                                                          & 0.7112          & 0.7044          & 0.7078          & 0.7040          \\
SCI (2019)                                                                                    & 0.8095          & 0.8082          & 0.8089          & 0.8076          \\
SCI (2020-24)                                                                                 & 0.8255          & 0.8570          & 0.8410          & 0.8482          \\
SCI+HCs (2019)                                                                                & 0.8583          & 0.8533          & 0.8558          & 0.8550          \\
HCs (2020-24)                                                                                 & \textbf{0.8725} & \textbf{0.8724} & \textbf{0.8724} & \textbf{0.8688} \\
SCI+HCs+Tribunal (2019)                                                                       & 0.8653          & 0.8624          & 0.8639          & 0.8629          \\
Tribunal (2020-24)                                                                            & 0.8625          & 0.8570          & 0.8598          & 0.8595          \\
\begin{tabular}[c]{@{}l@{}}SCI+HCs+Tribunal+\\ DailyOrders+DistrictCourts (2019)\end{tabular} & 0.8640          & 0.8601          & 0.8620          & 0.8610          \\
Daily\_orders (2020-24)                                                                       & 0.8282          & 0.8473          & 0.8376          & 0.8363          \\ \hline
\end{tabular}%
}
\caption{Judgment prediction results on the binary task across different court cases and temporal test cases, with models trained on SCI + HCs + Tribunal data from \texttt{\named} single-split data. The best results are highlighted in bold.}
\label{tab:binary-judgment-results-single-SCI-HCs-Tribunal}
\end{table}
\begin{table}[ht]
\centering
\resizebox{\columnwidth}{!}{%
\begin{tabular}{lrrrr}
 \hline
\multicolumn{1}{l}{\textbf{Test Data}} &
  \multicolumn{1}{r}{\textbf{\begin{tabular}[c]{@{}c@{}}Macro \\ Precision\end{tabular}}} &
  \multicolumn{1}{r}{\textbf{\begin{tabular}[c]{@{}c@{}}Macro \\ Recall\end{tabular}}} &
  \multicolumn{1}{r}{\textbf{\begin{tabular}[c]{@{}c@{}}Macro \\ F1\end{tabular}}} &
  \multicolumn{1}{r}{\textbf{Accuracy}} \\ \hline
\multicolumn{5}{c}{\textbf{InLegalBert}}                                                                                                                              \\ \hline
ILDC                                                                                          & 0.7209          & 0.7169          & 0.7189          & 0.7172          \\
SCI (2019)                                                                                    & 0.8261          & 0.8255          & 0.8258          & 0.8258          \\
SCI (2020-24)                                                                                 & 0.8515          & 0.8588          & 0.8552          & 0.8720          \\
SCI+HCs (2019)                                                                                & 0.8739          & 0.8735          & 0.8737          & 0.8739          \\
HCs (2020-24)                                                                                 & \textbf{0.8940} & \textbf{0.8943} & \textbf{0.8942} & 0.8945          \\
SCI+HCs+Tribunal (2019)                                                                       & 0.8637          & 0.8634          & 0.8635          & 0.8635          \\
Tribunal (2020-24)                                                                            & 0.8308          & 0.8249          & 0.8278          & 0.8277          \\
\begin{tabular}[c]{@{}l@{}}SCI+HCs+Tribunal+\\ DailyOrders+DistrictCourts (2019)\end{tabular} & 0.8722          & 0.8718          & 0.8720          & 0.8720          \\
Daily\_orders (2020-24)                                                                       & 0.8897          & 0.8869          & 0.8883          & \textbf{0.8955} \\ \hline
\multicolumn{5}{c}{\textbf{InCaseLaw}}                                                                                                                                \\ \hline
ILDC                                                                                          & 0.7347          & 0.7335          & 0.7341          & 0.7337          \\
SCI (2019)                                                                                    & 0.8271          & 0.8272          & 0.8271          & 0.8272          \\
SCI (2020-24)                                                                                 & 0.8449          & 0.8579          & 0.8513          & 0.8670          \\
SCI+HCs (2019)                                                                                & 0.8585          & 0.8570          & 0.8578          & 0.8579          \\
HCs (2020-24)                                                                                 & \textbf{0.8891} & \textbf{0.8898} & \textbf{0.8895} & 0.8898          \\
SCI+HCs+Tribunal (2019)                                                                       & 0.8544          & 0.8521          & 0.8532          & 0.8526          \\
Tribunal (2020-24)                                                                            & 0.8160          & 0.7939          & 0.8048          & 0.8001          \\
\begin{tabular}[c]{@{}l@{}}SCI+HCs+Tribunal+\\ DailyOrders+DistrictCourts (2019)\end{tabular} & 0.8573          & 0.8553          & 0.8563          & 0.8559          \\
Daily\_orders (2020-24)                                                                       & 0.8868          & 0.8795          & 0.8831          & \textbf{0.8911} \\ \hline
\multicolumn{5}{c}{\textbf{XLNet Large}}                                                                                                                              \\ \hline
ILDC                                                                                          & 0.6851          & 0.6850          & 0.6851          & 0.6849          \\
SCI (2019)                                                                                    & 0.8150          & 0.8137          & 0.8143          & 0.8142          \\
SCI (2020-24)                                                                                 & 0.8507          & 0.8562          & 0.8535          & 0.8709          \\
SCI+HCs (2019)                                                                                & 0.8590          & 0.8585          & 0.8588          & 0.8590          \\
HCs (2020-24)                                                                                 & \textbf{0.8848} & 0.8863          & 0.8856          & \textbf{0.8851} \\ 
SCI+HCs+Tribunal (2019)                                                                       & 0.8580          & 0.8576          & 0.8578          & 0.8577          \\
Tribunal (2020-24)                                                                            & 0.8180          & 0.8053          & 0.8116          & 0.8098          \\ 
\begin{tabular}[c]{@{}l@{}}SCI+HCs+Tribunal+\\ DailyOrders+DistrictCourts (2019)\end{tabular} & 0.8660          & 0.8655          & 0.8657          & 0.8657          \\
Daily\_orders (2020-24)                                                                       & 0.8832          & \textbf{0.8904} & \textbf{0.8868} & 0.8924       \\   \hline
\end{tabular}%
}
\caption{Judgment prediction results on the binary task across different court cases and temporal test cases, with models trained on SCI + HCs + Tribunal + Daily Orders and District Court data from \texttt{\named} multi-split data. The best results are highlighted in bold.}
\label{tab:binary-judgment-results-multi}
\end{table}


\begin{table}[t]
\centering
\resizebox{\columnwidth}{!}{%
\begin{tabular}{lrrrr}
\hline
\multicolumn{1}{l}{\textbf{Test Data}} &
  \multicolumn{1}{r}{\textbf{\begin{tabular}[c]{@{}c@{}}Macro \\ Precision\end{tabular}}} &
  \multicolumn{1}{r}{\textbf{\begin{tabular}[c]{@{}c@{}}Macro \\ Recall\end{tabular}}} &
  \multicolumn{1}{r}{\textbf{\begin{tabular}[c]{@{}c@{}}Macro \\ F1\end{tabular}}} &
  \multicolumn{1}{r}{\textbf{Accuracy}} \\ \hline
\multicolumn{5}{c}{\textbf{InLegalBert}}                                                                                                                              \\ \hline
ILDC                                                                                          & 0.7492          & 0.7351          & 0.7421          & 0.7357          \\
SCI (2019)                                                                                    & 0.8532          & 0.8437          & 0.8484          & 0.8451          \\
SCI (2020-24)                                                                                 & \textbf{0.9102} & 0.8798          & \textbf{0.8947} & \textbf{0.9095} \\
SCI+HCs (2019)                                                                                & 0.8822          & 0.8814          & 0.8818          & 0.8799          \\
HCs (2020-24)                                                                                 & 0.8908          & \textbf{0.8869} & 0.8888          & 0.8891          \\
SCI+HCs+Tribunal (2019)                                                                       & 0.8785          & 0.8744          & 0.8764          & 0.8737          \\
Tribunal (2020-24)                                                                            & 0.8275          & 0.8194          & 0.8234          & 0.8142          \\
\begin{tabular}[c]{@{}l@{}}SCI+HCs+Tribunal+\\ DailyOrders+DistrictCourts (2019)\end{tabular} & 0.8544          & 0.8493          & 0.8519          & 0.8483          \\
Daily\_orders (2020-24)                                                                       & 0.8852          & 0.8686          & 0.8768          & 0.8855          \\ \hline
\multicolumn{5}{c}{\textbf{InCaseLaw}}                                                                                                                                \\ \hline
ILDC                                                                                          & 0.6856          & 0.6369          & 0.6604          & 0.6381          \\
SCI (2019)                                                                                    & 0.7965          & 0.7744          & 0.7852          & 0.7767          \\
SCI (2020-24)                                                                                 & \textbf{0.8673} & 0.8385          & \textbf{0.8526} & \textbf{0.8742} \\
SCI+HCs (2019)                                                                                & 0.8211          & 0.8152          & 0.8182          & 0.8124          \\
HCs (2020-24)                                                                                 & 0.8501          & \textbf{0.8446} & 0.8474          & 0.8477          \\
SCI+HCs+Tribunal (2019)                                                                       & 0.8262          & 0.8164          & 0.8213          & 0.8153          \\
Tribunal (2020-24)                                                                            & 0.8234          & 0.8194          & 0.8214          & 0.8154          \\
\begin{tabular}[c]{@{}l@{}}SCI+HCs+Tribunal+\\ DailyOrders+DistrictCourts (2019)\end{tabular} & 0.8259          & 0.8174          & 0.8216          & 0.8157          \\
Daily\_orders (2020-24)                                                                       & 0.8624          & 0.8379          & 0.8500          & 0.8610          \\ \hline
\multicolumn{5}{c}{\textbf{XLNet Large}}                                                                                                                              \\ \hline
ILDC                                                                                          & 0.7257          & 0.7107          & 0.7181          & 0.7113          \\
SCI (2019)                                                                                    & 0.8479          & 0.8356          & 0.8417          & 0.8371          \\
SCI (2020-24)                                                                                 & 0.8965          & 0.8794          & 0.8878          & 0.9034          \\
SCI+HCs (2019)                                                                                & 0.8686          & 0.8668          & 0.8677          & 0.8650          \\
HCs (2020-24)                                                                                 & \textbf{0.9065} & \textbf{0.9023} & \textbf{0.9044} & \textbf{0.9045} \\
SCI+HCs+Tribunal (2019)                                                                       & 0.8634          & 0.8571          & 0.8602          & 0.8562          \\
Tribunal (2020-24)                                                                            & 0.8255          & 0.8198          & 0.8227          & 0.8153          \\
\begin{tabular}[c]{@{}l@{}}SCI+HCs+Tribunal+\\ DailyOrders+DistrictCourts (2019)\end{tabular} & 0.8613          & 0.8557          & 0.8585          & 0.8544          \\
Daily\_orders (2020-24)                                                                       & 0.8813          & 0.8638          & 0.8725          & 0.8815          \\ \hline
\end{tabular}%
}
\caption{Judgment prediction results on the binary task across different court cases and temporal test cases, with models trained on SCI + HCs + Tribunal data from \texttt{\named} multi-split data. The best results are highlighted in bold.}
\label{tab:binary-judgment-results-multi-SCI-HCs-Tribunal}
\end{table}

\begin{table}[t]
\centering
\resizebox{\columnwidth}{!}{%
\begin{tabular}{lrrrr}
\hline
\multicolumn{1}{l}{\textbf{Test Data}} &
  \multicolumn{1}{r}{\textbf{\begin{tabular}[c]{@{}c@{}}Macro \\ Precision\end{tabular}}} &
  \multicolumn{1}{r}{\textbf{\begin{tabular}[c]{@{}c@{}}Macro \\ Recall\end{tabular}}} &
  \multicolumn{1}{r}{\textbf{\begin{tabular}[c]{@{}c@{}}Macro \\ F1\end{tabular}}} &
  \multicolumn{1}{r}{\textbf{Accuracy}} \\ \hline
\multicolumn{5}{c}{\textbf{InLegalBert}}                                                                                                                              \\ \hline
ILDC                                                                                          & 0.7544          & 0.7536          & 0.7540          & 0.7535          \\
SCI (2019)                                                                                    & 0.8309          & 0.8308          & 0.8308          & 0.8309          \\
SCI (2020-24)                                                                                 & 0.8742          & 0.8807          & 0.8774          & 0.8918          \\
SCI+HCs (2019)                                                                                & 0.8691          & 0.8692          & 0.8692          & 0.8693          \\
HCs (2020-24)                                                                                 & \textbf{0.8952} & \textbf{0.8953} & \textbf{0.8952} & \textbf{0.8956} \\
SCI+HCs+Tribunal (2019)                                                                       & 0.8542          & 0.8538          & 0.8540          & 0.8540          \\
Tribunal (2020-24)                                                                            & 0.8086          & 0.7849          & 0.7966          & 0.7914          \\
\begin{tabular}[c]{@{}l@{}}SCI+HCs+Tribunal+\\ DailyOrders+DistrictCourts (2019)\end{tabular} & 0.7499          & 0.7421          & 0.7460          & 0.7406          \\
Daily\_orders (2020-24)                                                                       & 0.8687          & 0.8684          & 0.8685          & 0.8767          \\ \hline
\multicolumn{5}{c}{\textbf{InCaseLaw}}                                                                                                                                \\ \hline
ILDC                                                                                          & 0.7513          & 0.7503          & 0.7508          & 0.7502          \\
SCI (2019)                                                                                    & 0.8327          & 0.8328          & 0.8328          & 0.8327          \\
SCI (2020-24)                                                                                 & 0.8555          & 0.8710          & 0.8632          & 0.8769          \\
SCI+HCs (2019)                                                                                & 0.8660          & 0.8652          & 0.8656          & 0.8658          \\
HCs (2020-24)                                                                                 & \textbf{0.8961} & \textbf{0.8969} & \textbf{0.8965} & \textbf{0.8967} \\
SCI+HCs+Tribunal (2019)                                                                       & 0.8511          & 0.8494          & 0.8503          & 0.8498          \\
Tribunal (2020-24)                                                                            & 0.7923          & 0.7599          & 0.7758          & 0.7679          \\
\begin{tabular}[c]{@{}l@{}}SCI+HCs+Tribunal+\\ DailyOrders+DistrictCourts (2019)\end{tabular} & 0.8497          & 0.8479          & 0.8488          & 0.8485          \\
Daily\_orders (2020-24)                                                                       & 0.8696          & 0.8721          & 0.8709          & 0.8784          \\ \hline
\multicolumn{5}{c}{\textbf{XLNet Large}}                                                                                                                              \\ \hline
ILDC                                                                                          & 0.7247          & 0.7245          & 0.7246          & 0.7245          \\
SCI (2019)                                                                                    & 0.8375          & 0.8368          & 0.8372          & 0.8371          \\
SCI (2020-24)                                                                                 & 0.8680          & 0.8792          & 0.8736          & 0.8874          \\
SCI+HCs (2019)                                                                                & 0.8788          & 0.8792          & 0.8790          & 0.8789          \\
HCs (2020-24)                                                                                 & \textbf{0.9101} & \textbf{0.9099} & \textbf{0.9100} & \textbf{0.9104} \\
SCI+HCs+Tribunal (2019)                                                                       & 0.8649          & 0.8650          & 0.8649          & 0.8649          \\
Tribunal (2020-24)                                                                            & 0.8114          & 0.8008          & 0.8060          & 0.8049          \\
\begin{tabular}[c]{@{}l@{}}SCI+HCs+Tribunal+\\ DailyOrders+DistrictCourts (2019)\end{tabular} & 0.8653          & 0.8654          & 0.8654          & 0.8654          \\
Daily\_orders (2020-24)                                                                       & 0.8684          & 0.8741          & 0.8713          & 0.8780          \\ \hline
\end{tabular}%
}
\caption{Judgment prediction results on the binary task across different court cases and temporal test cases, with models trained on SCI + HCs data from \texttt{\named} single split data. The best results are highlighted in bold.}
\label{tab:binary-judgment-results-single-SCI-HCs}
\end{table}

\begin{table}[t]
\centering
\resizebox{\columnwidth}{!}{%
\begin{tabular}{lrrrr}
\hline
\multicolumn{1}{l}{\textbf{Test Data}} &
  \multicolumn{1}{r}{\textbf{\begin{tabular}[c]{@{}c@{}}Macro \\ Precision\end{tabular}}} &
  \multicolumn{1}{r}{\textbf{\begin{tabular}[c]{@{}c@{}}Macro \\ Recall\end{tabular}}} &
  \multicolumn{1}{r}{\textbf{\begin{tabular}[c]{@{}c@{}}Macro \\ F1\end{tabular}}} &
  \multicolumn{1}{r}{\textbf{Accuracy}} \\ \hline
\multicolumn{5}{c}{\textbf{InLegalBert}}                                                                                                                              \\ \hline
ILDC                                                                                          & 0.7364          & 0.7078          & 0.7218          & 0.7086          \\
SCI (2019)                                                                                    & 0.8480          & 0.8323          & 0.8401          & 0.8341          \\
SCI (2020-24)                                                                                 & 0.8986          & 0.8614          & 0.8796          & 0.8968          \\
SCI+HCs (2019)                                                                                & 0.8689          & 0.8653          & 0.8671          & 0.8630          \\
HCs (2020-24)                                                                                 & \textbf{0.8949} & \textbf{0.8881} & \textbf{0.8915} & \textbf{0.8912} \\
SCI+HCs+Tribunal (2019)                                                                       & 0.8529          & 0.8474          & 0.8501          & 0.8465          \\
Tribunal (2020-24)                                                                            & 0.8069          & 0.8072          & 0.8071          & 0.8074          \\
\begin{tabular}[c]{@{}l@{}}SCI+HCs+Tribunal+\\ DailyOrders+DistrictCourts (2019)\end{tabular} & 0.7672          & 0.7406          & 0.7537          & 0.7381          \\
Daily\_orders (2020-24)                                                                       & 0.8779          & 0.8593          & 0.8685          & 0.8779          \\ \hline
\multicolumn{5}{c}{\textbf{InCaseLaw}}                                                                                                                                \\ \hline
ILDC                                                                                          & 0.6963          & 0.6675          & 0.6816          & 0.6684          \\
SCI (2019)                                                                                    & 0.8062          & 0.7930          & 0.7995          & 0.7947          \\
SCI (2020-24)                                                                                 & 0.8652          & 0.8516          & 0.8584          & \textbf{0.8780} \\
SCI+HCs (2019)                                                                                & 0.8315          & 0.8265          & 0.8290          & 0.8239          \\
HCs (2020-24)                                                                                 & \textbf{0.8678} & \textbf{0.8584} & \textbf{0.8631} & 0.8624          \\
SCI+HCs+Tribunal (2019)                                                                       & 0.8174          & 0.8112          & 0.8143          & 0.8102          \\
Tribunal (2020-24)                                                                            & 0.7831          & 0.7828          & 0.7829          & 0.7836          \\
\begin{tabular}[c]{@{}l@{}}SCI+HCs+Tribunal+\\ DailyOrders+DistrictCourts (2019)\end{tabular} & 0.8172          & 0.8118          & 0.8145          & 0.8104          \\
Daily\_orders (2020-24)                                                                       & 0.8583          & 0.8304          & 0.8441          & 0.8555          \\ \hline
\multicolumn{5}{c}{\textbf{XLNet Large}}                                                                                                                              \\ \hline
ILDC                                                                                          & 0.7265          & 0.7181          & 0.7223          & 0.7185          \\
SCI (2019)                                                                                    & 0.8412          & 0.8292          & 0.8351          & 0.8307          \\
SCI (2020-24)                                                                                 & \textbf{0.9118} & 0.8910          & 0.9013          & \textbf{0.9150} \\
SCI+HCs (2019)                                                                                & 0.8645          & 0.8624          & 0.8635          & 0.8605          \\
HCs (2020-24)                                                                                 & 0.9060          & \textbf{0.9014} & \textbf{0.9037} & 0.9037          \\
SCI+HCs+Tribunal (2019)                                                                       & 0.8505          & 0.8466          & 0.8485          & 0.8459          \\
Tribunal (2020-24)                                                                            & 0.8159          & 0.8166          & 0.8162          & 0.8164          \\
\begin{tabular}[c]{@{}l@{}}SCI+HCs+Tribunal+\\ DailyOrders+DistrictCourts (2019)\end{tabular} & 0.8506          & 0.8471          & 0.8488          & 0.8462          \\
Daily\_orders (2020-24)                                                                       & 0.8783          & 0.8606          & 0.8693          & 0.8787          \\ \hline
\end{tabular}%
}
\caption{Judgment prediction results on the binary task across different court cases and temporal test cases, with models trained on SCI + HCs data from \texttt{\named} multi-split data. The best results are highlighted in bold.}
\label{tab:binary-judgment-results-multi-SCI-HCs}
\end{table}

\begin{table}[t]
\centering
\resizebox{\columnwidth}{!}{%
\begin{tabular}{lrrrr}
\hline
\multicolumn{1}{l}{\textbf{Test Data}} &
  \multicolumn{1}{r}{\textbf{\begin{tabular}[c]{@{}c@{}}Macro \\ Precision\end{tabular}}} &
  \multicolumn{1}{r}{\textbf{\begin{tabular}[c]{@{}c@{}}Macro \\ Recall\end{tabular}}} &
  \multicolumn{1}{r}{\textbf{\begin{tabular}[c]{@{}c@{}}Macro \\ F1\end{tabular}}} &
  \multicolumn{1}{r}{\textbf{Accuracy}} \\ \hline
\multicolumn{5}{c}{\textbf{InLegalBert}}                                                                                                                              \\ \hline
ILDC                                                                                          & 0.7147          & 0.7145          & 0.7146          & 0.7146          \\
SCI (2019)                                                                                    & 0.8212          & 0.8187          & 0.8200          & 0.8194          \\
SCI (2020-24)                                                                                 & \textbf{0.8983} & \textbf{0.8532} & \textbf{0.8752} & \textbf{0.8929}  \\
SCI+HCs (2019)                                                                                & 0.7691          & 0.7664          & 0.7677          & 0.7642          \\
HCs (2020-24)                                                                                 & 0.7919          & 0.7585          & 0.7748          & 0.7683          \\
SCI+HCs+Tribunal (2019)                                                                       & 0.7514          & 0.7447          & 0.7481          & 0.7437          \\
Tribunal (2020-24)                                                                            & 0.7300          & 0.6655          & 0.6963          & 0.6514          \\
\begin{tabular}[c]{@{}l@{}}SCI+HCs+Tribunal+\\ DailyOrders+DistrictCourts (2019)\end{tabular} & 0.7499          & 0.7421          & 0.7460          & 0.7406          \\
Daily\_orders (2020-24)                                                                       & 0.7849          & 0.7389          & 0.7612          & 0.7814          \\ \hline
\multicolumn{5}{c}{\textbf{InCaseLaw}}                                                                                                                                \\ \hline
ILDC                                                                                          & 0.7067          & 0.7066          & 0.7066          & 0.7067          \\
SCI (2019)                                                                                    & 0.8093          & 0.8083          & 0.8088          & 0.8088          \\
SCI (2020-24)                                                                                 & \textbf{0.8782} & 0.8394          & 0.8584          & \textbf{0.8791} \\
SCI+HCs (2019)                                                                                & 0.7596          & 0.7578          & 0.7587          & 0.7559          \\
HCs (2020-24)                                                                                 & 0.8678          & \textbf{0.8584} & \textbf{0.8631} & 0.8624          \\
SCI+HCs+Tribunal (2019)                                                                       & 0.7420          & 0.7369          & 0.7394          & 0.7359          \\
Tribunal (2020-24)                                                                            & 0.7031          & 0.6493          & 0.6751          & 0.6355          \\
\begin{tabular}[c]{@{}l@{}}SCI+HCs+Tribunal+\\ DailyOrders+DistrictCourts (2019)\end{tabular} & 0.7397          & 0.7340          & 0.7368          & 0.7324          \\
Daily\_orders (2020-24)                                                                       & 0.7826          & 0.7356          & 0.7584          & 0.7790          \\ \hline
\multicolumn{5}{c}{\textbf{XLNet Large}}                                                                                                                              \\ \hline
ILDC                                                                                          & 0.7356          & 0.7342          & 0.7349          & 0.7343          \\
SCI (2019)                                                                                    & 0.8495          & 0.8488          & 0.8492          & 0.8491          \\
SCI (2020-24)                                                                                 & \textbf{0.9185} & \textbf{0.8961} & \textbf{0.9071} & \textbf{0.9200} \\
SCI+HCs (2019)                                                                                & 0.8064          & 0.8063          & 0.8063          & 0.8052          \\
HCs (2020-24)                                                                                 & 0.8342          & 0.8288          & 0.8315          & 0.8320          \\
SCI+HCs+Tribunal (2019)                                                                       & 0.7936          & 0.7912          & 0.7924          & 0.7906          \\
Tribunal (2020-24)                                                                            & 0.7810          & 0.7629          & 0.7718          & 0.7555          \\
\begin{tabular}[c]{@{}l@{}}SCI+HCs+Tribunal+\\ DailyOrders+DistrictCourts (2019)\end{tabular} & 0.7916          & 0.7894          & 0.7905          & 0.7887          \\
Daily\_orders (2020-24)                                                                       & 0.8224          & 0.8167          & 0.8195          & 0.8319          \\ \hline
\end{tabular}%
}
\caption{Judgment prediction results on the binary task across different court cases and temporal test cases, with models trained on SCI data from \texttt{\named} single split data. The best results are highlighted in bold.}
\label{tab:binary-judgment-results-single-SCI}
\end{table}

\begin{table}[t]
\centering
\resizebox{\columnwidth}{!}{%
\begin{tabular}{lrrrr}
\hline
\multicolumn{1}{l}{\textbf{Test Data}} &
  \multicolumn{1}{r}{\textbf{\begin{tabular}[c]{@{}c@{}}Macro \\ Precision\end{tabular}}} &
  \multicolumn{1}{r}{\textbf{\begin{tabular}[c]{@{}c@{}}Macro \\ Recall\end{tabular}}} &
  \multicolumn{1}{r}{\textbf{\begin{tabular}[c]{@{}c@{}}Macro \\ F1\end{tabular}}} &
  \multicolumn{1}{r}{\textbf{Accuracy}} \\ \hline
\multicolumn{5}{c}{\textbf{InLegalBert}}                                                                                                                              \\ \hline
ILDC                                                                                          & 0.7216          & 0.6866          & 0.7037          & 0.6875          \\
SCI (2019)                                                                                    & 0.8370          & 0.8230          & 0.8299          & 0.8246          \\
SCI (2020-24)                                                                                 & \textbf{0.9107} & \textbf{0.8546} & \textbf{0.8818} & \textbf{0.8979} \\
SCI+HCs (2019)                                                                                & 0.7781          & 0.7623          & 0.7701          & 0.7577          \\
HCs (2020-24)                                                                                 & 0.8051          & 0.7709          & 0.7876          & 0.7806          \\
SCI+HCs+Tribunal (2019)                                                                       & 0.7678          & 0.7427          & 0.7550          & 0.7407          \\
Tribunal (2020-24)                                                                            & 0.7354          & 0.6537          & 0.6921          & 0.6380          \\
\begin{tabular}[c]{@{}l@{}}SCI+HCs+Tribunal+\\ DailyOrders+DistrictCourts (2019)\end{tabular} & 0.7672          & 0.7406          & 0.7537          & 0.7381          \\
Daily\_orders (2020-24)                                                                       & 0.8241          & 0.7509          & 0.7858          & 0.8003          \\ \hline
\multicolumn{5}{c}{\textbf{InCaseLaw}}                                                                                                                                \\ \hline
ILDC                                                                                          & 0.7134          & 0.6881          & 0.7005          & 0.6889          \\
SCI (2019)                                                                                    & 0.8385          & 0.8265          & 0.8325          & 0.8280          \\
SCI (2020-24)                                                                                 & \textbf{0.9042} & 0.8535          & \textbf{0.8781} & \textbf{0.8951} \\
SCI+HCs (2019)                                                                                & 0.7696          & 0.7546          & 0.7620          & 0.7501          \\
HCs (2020-24)                                                                                 & 0.8678          & \textbf{0.8584} & 0.8631          & 0.8624          \\
SCI+HCs+Tribunal (2019)                                                                       & 0.7631          & 0.7391          & 0.7509          & 0.7371          \\
Tribunal (2020-24)                                                                            & 0.7272          & 0.6485          & 0.6856          & 0.6328          \\
\begin{tabular}[c]{@{}l@{}}SCI+HCs+Tribunal+\\ DailyOrders+DistrictCourts (2019)\end{tabular} & 0.7607          & 0.7367          & 0.7485          & 0.7337          \\
Daily\_orders (2020-24)                                                                       & 0.8140          & 0.7386          & 0.7744          & 0.7903          \\ \hline
\multicolumn{5}{c}{\textbf{XLNet Large}}                                                                                                                              \\ \hline
ILDC                                                                                          & 0.7229          & 0.7020          & 0.7123          & 0.7027          \\
SCI (2019)                                                                                    & 0.8851          & 0.8776          & 0.8813          & 0.8787          \\
SCI (2020-24)                                                                                 & \textbf{0.9187} & \textbf{0.8787} & \textbf{0.8982} & \textbf{0.9123} \\
SCI+HCs (2019)                                                                                & 0.8026          & 0.7921          & 0.7973          & 0.7884          \\
HCs (2020-24)                                                                                 & 0.8282          & 0.8097          & 0.8189          & 0.8162          \\
SCI+HCs+Tribunal (2019)                                                                       & 0.7913          & 0.7726          & 0.7818          & 0.7710          \\
Tribunal (2020-24)                                                                            & 0.7748          & 0.7402          & 0.7571          & 0.7303          \\
\begin{tabular}[c]{@{}l@{}}SCI+HCs+Tribunal+\\ DailyOrders+DistrictCourts (2019)\end{tabular} & 0.7901          & 0.7717          & 0.7808          & 0.7697          \\
Daily\_orders (2020-24)                                                                       & 0.8300          & 0.7825          & 0.8056          & 0.8201          \\ \hline
\end{tabular}%
}
\caption{Judgment prediction results on the binary task across different court cases and temporal test cases, with models trained on SCI data from \texttt{\named} multi-split data. The best results are highlighted in bold.}
\label{tab:binary-judgment-results-multi-SCI}
\end{table}

\begin{table}[t]
\centering
\resizebox{\columnwidth}{!}{%
\begin{tabular}{lrrrr}
\hline
\multicolumn{1}{l}{\textbf{Test Data}} &
  \multicolumn{1}{r}{\textbf{\begin{tabular}[c]{@{}c@{}}Macro \\ Precision\end{tabular}}} &
  \multicolumn{1}{r}{\textbf{\begin{tabular}[c]{@{}c@{}}Macro \\ Recall\end{tabular}}} &
  \multicolumn{1}{r}{\textbf{\begin{tabular}[c]{@{}c@{}}Macro \\ F1\end{tabular}}} &
  \multicolumn{1}{r}{\textbf{Accuracy}} \\ \hline
\multicolumn{5}{c}{\textbf{InLegalBert}}                                                                                                                              \\ \hline
ILDC                                                                                          & 0.7447          & 0.7437          & 0.7442          & 0.7436          \\
SCI (2019)                                                                                    & 0.7559          & 0.7516          & 0.7538          & 0.7527          \\
SCI (2020-24)                                                                                 & \textbf{0.8259} & \textbf{0.7305} & \textbf{0.7753} & \textbf{0.8113} \\
SCI+HCs (2019)                                                                                & 0.6810          & 0.6657          & 0.6733          & 0.6604          \\
HCs (2020-24)                                                                                 & 0.6766          & 0.6155          & 0.6446          & 0.6338          \\
SCI+HCs+Tribunal (2019)                                                                       & 0.6562          & 0.6384          & 0.6472          & 0.6362          \\
Tribunal (2020-24)                                                                            & 0.6446          & 0.5566          & 0.5974          & 0.5363          \\
\begin{tabular}[c]{@{}l@{}}SCI+HCs+Tribunal+\\ DailyOrders+DistrictCourts (2019)\end{tabular} & 0.6793          & 0.6722          & 0.6757          & 0.6737          \\
Daily\_orders (2020-24)                                                                       & 0.6865          & 0.6033          & 0.6422          & 0.6829          \\ \hline
\multicolumn{5}{c}{\textbf{InCaseLaw}}                                                                                                                                \\ \hline
ILDC                                                                                          & 0.7451          & 0.7380          & 0.7415          & 0.7376          \\
SCI (2019)                                                                                    & 0.7367          & 0.7347          & 0.7357          & 0.7354          \\
SCI (2020-24)                                                                                 & \textbf{0.7876} & \textbf{0.6971} & \textbf{0.7396} & \textbf{0.7848} \\
SCI+HCs (2019)                                                                                & 0.6573          & 0.6460          & 0.6516          & 0.6411          \\
HCs (2020-24)                                                                                 & 0.6306          & 0.5864          & 0.6077          & 0.6046          \\
SCI+HCs+Tribunal (2019)                                                                       & 0.6341          & 0.6188          & 0.6264          & 0.6167          \\
Tribunal (2020-24)                                                                            & 0.6362          & 0.5480          & 0.5888          & 0.5271          \\
\begin{tabular}[c]{@{}l@{}}SCI+HCs+Tribunal+\\ DailyOrders+DistrictCourts (2019)\end{tabular} & 0.6333          & 0.6158          & 0.6244          & 0.6123          \\
Daily\_orders (2020-24)                                                                       & 0.6648          & 0.5959          & 0.6285          & 0.6736          \\ \hline
\multicolumn{5}{c}{\textbf{XLNet Large}}                                                                                                                              \\ \hline
ILDC                                                                                          & 0.7429          & 0.7361          & 0.7395          & 0.7357          \\
SCI (2019)                                                                                    & 0.7577          & 0.7575          & 0.7576          & 0.7577          \\
SCI (2020-24)                                                                                 & \textbf{0.7963} & \textbf{0.7855} & \textbf{0.7909} & \textbf{0.8201} \\
SCI+HCs (2019)                                                                                & 0.7114          & 0.7114          & 0.7114          & 0.7118          \\
HCs (2020-24)                                                                                 & 0.6982          & 0.6988          & 0.6985          & 0.6990          \\
SCI+HCs+Tribunal (2019)                                                                       & 0.6897          & 0.6897          & 0.6897          & 0.6897          \\
Tribunal (2020-24)                                                                            & 0.6613          & 0.6515          & 0.6564          & 0.6446          \\
\begin{tabular}[c]{@{}l@{}}SCI+HCs+Tribunal+\\ DailyOrders+DistrictCourts (2019)\end{tabular} & 0.6895          & 0.6895          & 0.6895          & 0.6894          \\
Daily\_orders (2020-24)                                                                       & 0.7028          & 0.7064          & 0.7046          & 0.7199          \\ \hline
\end{tabular}%
}
\caption{Judgment prediction results on the binary task across different court cases and temporal test cases, with models trained on data from ILDC single data. The best results are highlighted in bold.}
\label{tab:binary-judgment-results-ILDC-single}
\end{table}

\begin{table}[t]
\centering
\resizebox{\columnwidth}{!}{%
\begin{tabular}{lrrrr}
\hline
\multicolumn{1}{l}{\textbf{Test Data}} &
  \multicolumn{1}{r}{\textbf{\begin{tabular}[c]{@{}c@{}}Macro \\ Precision\end{tabular}}} &
  \multicolumn{1}{r}{\textbf{\begin{tabular}[c]{@{}c@{}}Macro \\ Recall\end{tabular}}} &
  \multicolumn{1}{r}{\textbf{\begin{tabular}[c]{@{}c@{}}Macro \\ F1\end{tabular}}} &
  \multicolumn{1}{r}{\textbf{Accuracy}} \\ \hline
\multicolumn{5}{c}{\textbf{InLegalBert}}                                                                                                                              \\ \hline
ILDC                                                                                          & 0.7706          & 0.7650          & 0.7678          & 0.7647          \\
SCI (2019)                                                                                    & 0.7799          & 0.7735          & 0.7767          & 0.7721          \\
SCI (2020-24)                                                                                 & \textbf{0.7868} & \textbf{0.8231} & \textbf{0.8045} & \textbf{0.8046} \\
SCI+HCs (2019)                                                                                & 0.7035          & 0.6907          & 0.6971          & 0.6949          \\
HCs (2020-24)                                                                                 & 0.6551          & 0.6446          & 0.6498          & 0.6355          \\
SCI+HCs+Tribunal (2019)                                                                       & 0.6840          & 0.6770          & 0.6805          & 0.6782          \\
Tribunal (2020-24)                                                                            & 0.6463          & 0.6464          & 0.6463          & 0.6447          \\
\begin{tabular}[c]{@{}l@{}}SCI+HCs+Tribunal+\\ DailyOrders+DistrictCourts (2019)\end{tabular} & 0.6544          & 0.6337          & 0.6439          & 0.6308          \\
Daily\_orders (2020-24)                                                                       & 0.5982          & 0.5918          & 0.5950          & 0.5428          \\ \hline
\multicolumn{5}{c}{\textbf{InCaseLaw}}                                                                                                                                \\ \hline
ILDC                                                                                          & 0.7592          & 0.7461          & 0.7526          & 0.7456          \\
SCI (2019)                                                                                    & \textbf{0.7691} & 0.7622          & 0.7656          & 0.7608          \\
SCI (2020-24)                                                                                 & 0.7687          & \textbf{0.8068} & \textbf{0.7873} & \textbf{0.7765} \\
SCI+HCs (2019)                                                                                & 0.6908          & 0.6822          & 0.6865          & 0.6857          \\
HCs (2020-24)                                                                                 & 0.6367          & 0.6271          & 0.6319          & 0.6180          \\
SCI+HCs+Tribunal (2019)                                                                       & 0.6691          & 0.6651          & 0.6671          & 0.6661          \\
Tribunal (2020-24)                                                                            & 0.6242          & 0.6241          & 0.6242          & 0.6221          \\
\begin{tabular}[c]{@{}l@{}}SCI+HCs+Tribunal+\\ DailyOrders+DistrictCourts (2019)\end{tabular} & 0.6667          & 0.6628          & 0.6647          & 0.6642          \\
Daily\_orders (2020-24)                                                                       & 0.5842          & 0.5818          & 0.5830          & 0.5399          \\ \hline
\multicolumn{5}{c}{\textbf{XLNet Large}}                                                                                                                              \\ \hline
ILDC                                                                                          & 0.7873          & 0.7840          & 0.7856          & 0.7838          \\
SCI (2019)                                                                                    & \textbf{0.8104} & \textbf{0.7959} & \textbf{0.8031} & \textbf{0.7939} \\
SCI (2020-24)                                                                                 & 0.7483          & 0.7839          & 0.7657          & 0.7417          \\
SCI+HCs (2019)                                                                                & 0.7224          & 0.7023          & 0.7122          & 0.7074          \\
HCs (2020-24)                                                                                 & 0.6918          & 0.6677          & 0.6795          & 0.6556          \\
SCI+HCs+Tribunal (2019)                                                                       & 0.7100          & 0.6949          & 0.7023          & 0.6965          \\
Tribunal (2020-24)                                                                            & 0.6704          & 0.6559          & 0.6631          & 0.6631          \\
\begin{tabular}[c]{@{}l@{}}SCI+HCs+Tribunal+\\ DailyOrders+DistrictCourts (2019)\end{tabular} & 0.7096          & 0.6942          & 0.7018          & 0.6963          \\
Daily\_orders (2020-24)                                                                       & 0.6358          & 0.6247          & 0.6302          & 0.5709          \\ \hline
\end{tabular}%
}
\caption{Judgment prediction results on the binary task across different court cases and temporal test cases, with models trained on data from ILDC multi data. The best results are highlighted in bold.}
\label{tab:binary-judgment-results-ILDC-multi}
\end{table}

\begin{table}[ht]
\centering
\resizebox{\columnwidth}{!}{%
\begin{tabular}{llrrrrrr}
\toprule
\textbf{Models} & \textbf{Metric} & \textbf{Overall} & \textbf{Class 0} & \textbf{Class 1} & \textbf{Class 2} \\ \midrule
\multirow{3}{*}{InLegalBert} 
& Macro Precision & 0.6950 & 0.81 & 0.84 & 0.44 \\ 
& Macro Recall    & \textbf{0.5883} & 0.82 & 0.85 & 0.10 \\ 
& Macro F1        & \textbf{0.6062} & 0.81 & 0.85 & 0.16 \\ \midrule

\multirow{3}{*}{InCaseLaw} 
& Macro Precision & \textbf{0.6984} & 0.80 & 0.83 & 0.46 \\ 
& Macro Recall    & 0.5608 & 0.81 & 0.84 & 0.03 \\ 
& Macro F1        & 0.5653 & 0.81 & 0.84 & 0.05 \\ \midrule

\multirow{3}{*}{XLNet} 
& Macro Precision & 0.6853 & 0.80 & 0.84 & 0.42 \\ 
& Macro Recall    & 0.5800 & 0.81 & 0.84 & 0.08 \\ 
& Macro F1        & 0.5952 & 0.81 & 0.84 & 0.14 \\ 
\bottomrule
\end{tabular}
}
\caption{Judgment prediction results on the ternary task on SCI + HCs court cases. The best results are highlighted in bold.}
\label{tab:judgment-prediction-ternary-sci-HCs}
\end{table}


\begin{table}[ht]
\centering
\resizebox{\columnwidth}{!}{%
\begin{tabular}{llrrrrrr}
\toprule
\textbf{Models} & \textbf{Metric} & \textbf{Overall} & \textbf{Class 0} & \textbf{Class 1} & \textbf{Class 2} \\ \midrule
\multirow{4}{*}{InLegalBert} 
& Macro Precision & \textbf{0.5432} & 0.80 & 0.83 & 0.00 \\ 
& Macro Recall    & \textbf{0.5453} & 0.77 & 0.87 & 0.00 \\ 
& Macro F1        & \textbf{0.5440} & 0.79 & 0.85 & 0.00 \\ \midrule 

\multirow{4}{*}{InCaseLaw} 
& Macro Precision & 0.4957 & 0.72 & 0.77 & 0.00 \\ 
& Macro Recall    & 0.4981 & 0.69 & 0.81 & 0.00 \\ 
& Macro F1        & 0.4966 & 0.70 & 0.79 & 0.00 \\ \midrule

\multirow{4}{*}{XLNet} 
& Macro Precision & 0.5376 & 0.79 & 0.82 & 0.00 \\ 
& Macro Recall    & 0.5411 & 0.77 & 0.85 & 0.00 \\ 
& Macro F1        & 0.5392 & 0.78 & 0.84 & 0.00 \\ \bottomrule

\end{tabular}%
}
\caption{Judgment prediction results on the ternary task on SCI + HCs + Tribunals court cases. The best results are highlighted in bold.}
\label{tab:judgment-prediction-ternary-sci-hcs-tribunals}
\end{table}


\begin{table}[ht]
\centering
\resizebox{\columnwidth}{!}{%
\begin{tabular}{llrrrrrr}
\toprule
\textbf{Models} & \textbf{Metric} & \textbf{Overall} & \textbf{Class 0} & \textbf{Class 1} & \textbf{Class 2} \\ \midrule
\multirow{4}{*}{InLegalBert} 
& Macro Precision & \textbf{0.5401} & 0.77 & 0.85 & 0.00 \\ 
& Macro Recall    & \textbf{0.5476} & 0.82 & 0.83 & 0.00 \\ 
& Macro F1        & \textbf{0.5436} & 0.79 & 0.84 & 0.00 \\ \midrule

\multirow{4}{*}{InCaseLaw} 
& Macro Precision & 0.4516 & 0.63 & 0.73 & 0.00 \\ 
& Macro Recall    & 0.4564 & 0.64 & 0.72 & 0.00 \\ 
& Macro F1        & 0.4540 & 0.64 & 0.73 & 0.00 \\ \midrule

\multirow{4}{*}{XLNet} 
& Macro Precision & 0.5362 & 0.76 & 0.85 & 0.00 \\ 
& Macro Recall    & 0.5446 & 0.82 & 0.81 & 0.00 \\ 
& Macro F1        & 0.5399 & 0.79 & 0.83 & 0.00 \\ \bottomrule

\end{tabular}%
}
\caption{Judgment prediction results on the ternary task on SCI + HCs + Tribunals + Daily Orders + District Court cases. The best results are highlighted in bold.}
\label{tab:judgment-prediction-ternary-sci-hcs-tribunals-dailyorders}
\end{table}

\begin{figure}[t]
    \centering
    \includegraphics[width=\columnwidth]{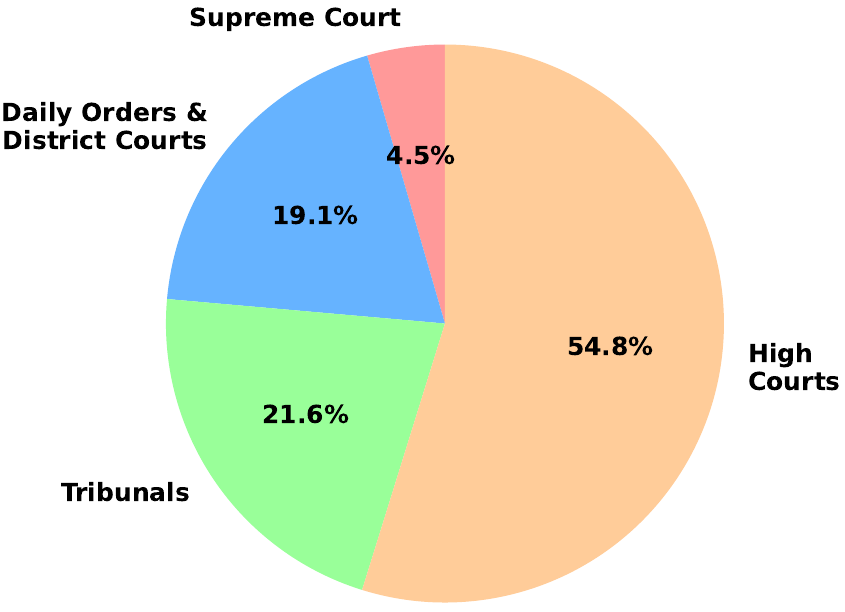}
    \caption{Distribution of cases in different courts.}
    \label{fig:court-distribution}
\end{figure}
\section{Preprocessing}
The preprocessing of the \texttt{\named} dataset involved several critical steps to ensure the data's quality and relevance. Given the unstructured nature of the documents, which varied in format and size, we faced challenges such as spelling errors and inconsistencies. To mitigate these, we used regular expressions to remove noisy text and meta-information, including case numbers, titles, judge names, petitioners, respondents, and dates. We identified key sections using specific terms like `ORDER,' `JUDGMENT,' and `JUDGEMENT' to isolate the essential content and filter out irrelevant details.

We further refined the dataset by removing cases without clear outcomes or insufficient information. To maintain consistency and manageability, we excluded extremely short cases (less than 50 words) and excessively long ones (more than 32,000 words). Additionally, tokens with characters repeated more than twice consecutively were removed to clean up text errors. This meticulous refinement process reduced the number of cases to 11,25,604, retaining only the most relevant cases, as detailed in Table \ref{tab:before-and-after}, which compares the number of cases before and after preprocessing, categorized by court type.

\subsection{Label Making}
After filtering, the documents were labeled using keywords indicative of positive outcomes like ``approved," ``allowed," and ``granted," or negative outcomes such as ``rejected," ``disapproved," and ``dismissed." This labeling process helped classify the cases into likely acceptance or rejection categories. We categorized the cases into two groups: single-labeled cases, where all appeals had the same outcome or only a single appeal was filed, marked as 0 (rejection) or 1 (acceptance), and multi-labeled judgments, where at least one appeal was accepted, indicating partial acceptance, marked as 2.

To ensure accurate labeling, we focused on the last 750 words of each document, typically where decisions are summarized. Special attention was given to a context window around key terms like ``appeal," ``petition," or ``case" to accurately determine the judgment nature. For example, phrases like ``Appeals Allowed" or ``The appeal is granted" indicated a positive outcome, while ``The appeal has no proper evidence and hence we reject it" indicated rejection. We also looked for indicators like ``partly" to identify multi-labeled judgments for cases with partial approvals. In instances where the judgment was phrased negatively (e.g., ``No appeal is allowed"), we used a label-flipping strategy if negation words like ``no" or ``not" were found close to key terms, ensuring the labels accurately reflected the judgment's intent.

This meticulous approach to labeling, focusing on the judgment's context and nuanced expression, enhances the reliability of our dataset and prepares it for effective training of judgment prediction models.

\begin{figure*}[t] 
    \centering 
    \includegraphics[width=\linewidth]{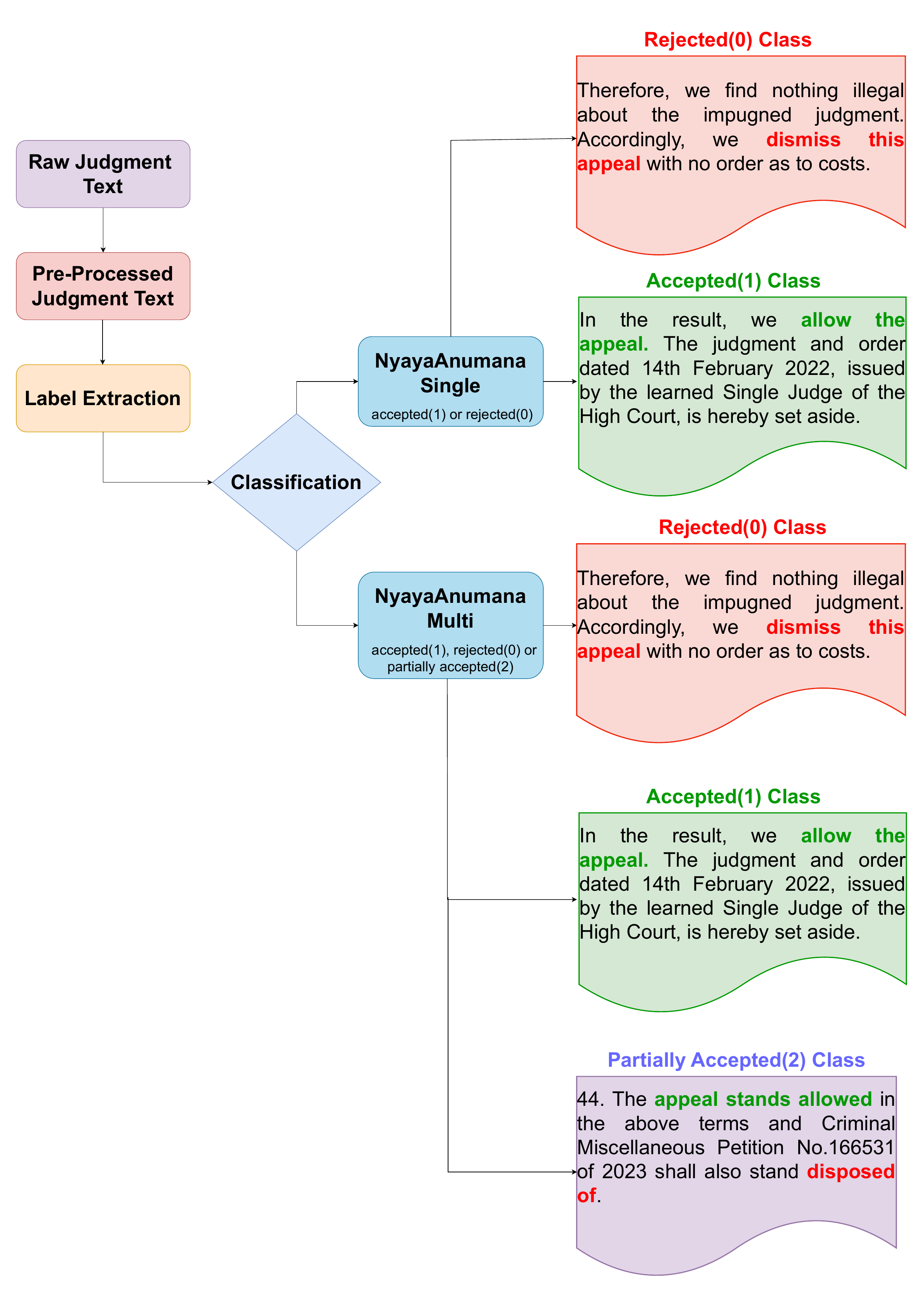} 
  
    \caption{Illustration of the LJP Task Framework.}
    \label{fig:task-framework} 
\end{figure*}

\begin{table}[ht]
\centering
\resizebox{\columnwidth}{!}{%
\begin{tabular}{lrrr}
\hline
\textbf{Court-wise} &
  \textbf{Raw files} &
  \textbf{\begin{tabular}[c]{@{}c@{}}FIles After \\ Preprocessing\end{tabular}} &
  \textbf{\begin{tabular}[c]{@{}c@{}}FIles After \\ Lebeling\end{tabular}} \\ \hline
\textbf{Supreme Court}  & 55,928   & 54,831   & 40,562  \\
\textbf{High Court}     & 13,24,373 & 9,77,849  & 4,82,295 \\
\textbf{Tribunal Court} & 4,77,397  & 3,18,681  & 1,86,671 \\
\textbf{\begin{tabular}[c]{@{}l@{}}Daily Orders and\\ District Courts\end{tabular}}    & 4,24,439  & 3,12,480  & 92,771 \\ 
\hline
\textbf{Total}          & 22,82,137 & 16,63,841 & 8,02,299
\\ \hline
\end{tabular}%
}
\caption{Number of cases before and after preprocessing, by court type.}
\label{tab:before-and-after}
\end{table}


\begin{table}[t]
\centering
\resizebox{0.8\columnwidth}{!}{%
\begin{tabular}{lccccc}
\toprule
 &
  \multicolumn{5}{c}{\textbf{Rating Score}} \\ \midrule
 
 &
  \multicolumn{1}{c}{\textbf{1}} &
  \multicolumn{1}{c}{\textbf{2}} &
  \multicolumn{1}{c}{\textbf{3}} &
  \multicolumn{1}{c}{\textbf{4}} &
  \textbf{5} \\ \cline{2-6} 
 
\multirow{-2}{*}{\textbf{Generative Models}} &
  \multicolumn{5}{c}{\textbf{PredEx}} \\ \midrule
LLaMa-2-7B &
  \multicolumn{1}{c}{{ 2}} &
  \multicolumn{1}{c}{{ 11}} &
  \multicolumn{1}{c}{{ 22}} &
  \multicolumn{1}{c}{{ 12}} &
  { 3} \\ 
\begin{tabular}[c]{@{}l@{}}LLaMa-2 SFT\end{tabular} &
  \multicolumn{1}{c}{{ 5}} &
  \multicolumn{1}{c}{13} &
  \multicolumn{1}{c}{18} &
  \multicolumn{1}{c}{13} &
  1 \\ 

\begin{tabular}[c]{@{}l@{}}LLaMa-2 CPT\end{tabular} &
  \multicolumn{1}{c}{{ 2}} &
  \multicolumn{1}{c}{2} &
  \multicolumn{1}{c}{27} &
  \multicolumn{1}{c}{19} &
  0 \\ 

\begin{tabular}[c]{@{}l@{}}\textbf{\namel} \\\textbf{CPT+SFT}\end{tabular} &
  \multicolumn{1}{c}{{ 0}} &
  \multicolumn{1}{c}{0} &
  \multicolumn{1}{c}{23} &
  \multicolumn{1}{c}{27} &
  0 \\ \midrule
 &
  \multicolumn{5}{c}{\textbf{ILDC\_expert}} \\ \midrule
LLaMa-2-7B &
  \multicolumn{1}{c}{0} &
  \multicolumn{1}{c}{9} &
  \multicolumn{1}{c}{22} &
  \multicolumn{1}{c}{21} &
  2 \\ 
\begin{tabular}[c]{@{}l@{}}LLaMa-2 SFT\end{tabular} &
  \multicolumn{1}{c}{2} &
  \multicolumn{1}{c}{3} &
  \multicolumn{1}{c}{16} &
  \multicolumn{1}{c}{24} &
  9 \\ 

  \begin{tabular}[c]{@{}l@{}}LLaMa-2 CPT\end{tabular} &
  \multicolumn{1}{c}{{1}} &
  \multicolumn{1}{c}{3} &
  \multicolumn{1}{c}{25} &
  \multicolumn{1}{c}{23} &
  2 \\ 

\begin{tabular}[c]{@{}l@{}}\textbf{\namel} \\\textbf{CPT+SFT}\end{tabular} &
  \multicolumn{1}{c}{{ 0}} &
  \multicolumn{1}{c}{0} &
  \multicolumn{1}{c}{22} &
  \multicolumn{1}{c}{28} &
  4 \\ \bottomrule
\end{tabular}%
}
\caption{Distribution of Expert Rating Scores for Generative Models on PredEx and ILDC\_expert Data.}
\label{expert-scores}
\end{table}